\documentclass{article}

    \PassOptionsToPackage{numbers, compress}{natbib}

\usepackage[final]{neurips_2024}

\usepackage[utf8]{inputenc} 
\usepackage[T1]{fontenc}    
\usepackage{hyperref}       
\usepackage{url}            
\usepackage{booktabs}       
\usepackage{amsfonts}       
\usepackage{nicefrac}       
\usepackage{microtype}      
\usepackage{xcolor}         
\definecolor{umn_maroon}{RGB}{122, 0, 25}
\definecolor{myred}{HTML}{D62728}
\definecolor{myblue}{HTML}{1F77B4}
\definecolor{mygreen}{HTML}{00FF00}
\definecolor{darkpastelgreen}{rgb}{0.01, 0.75, 0.24}

\usepackage{graphicx,amsmath,amsfonts,amsthm,amscd,amssymb,bm,url,color,latexsym}
\usepackage{physics}
\allowdisplaybreaks
\usepackage[capitalize,nameinlink]{cleveref}

\usepackage{bbold}
\renewcommand{\mathbf}{\boldsymbol}

\newcommand{\mb}{\mathbf}
\newcommand{\mc}{\mathcal}

\newcommand{\bb}{\mathbb}

\newcommand{\set}[1]{\left\{ #1 \right\}}

\newcommand{\paren}{\pqty}

\DeclareMathOperator*{\argmin}{arg\,min}

\newcommand{\wh}{\widehat}

\newcommand{\ol}{\overline}

\newcommand{\T}{\intercal}

\usepackage{wrapfig}
\usepackage{enumitem}
\usepackage{algorithm}%
\usepackage{algpseudocode}

\usepackage{siunitx}
\usepackage{multirow}
\usepackage{tikz}
\usepackage{sidecap}
\usetikzlibrary{decorations.pathreplacing,calc}
\newcommand{\tikzmark}[1]{\tikz[overlay,remember picture] \node (#1) {};}

\newcommand*{\AddNote}[4]{%
    \begin{tikzpicture}[overlay, remember picture]
        \draw [decoration={brace,amplitude=0.5em},decorate,ultra thick,red]
            ($(#3)!(#1.north)!($(#3)-(0,1)$)$) --  
            ($(#3)!(#2.south)!($(#3)-(0,1)$)$)
                node [align=left, text width=2.5cm, pos=0.5, anchor=west] {#4};
    \end{tikzpicture}
}%
\title{DMPlug: A Plug-in Method for Solving Inverse Problems with Diffusion Models}

\author{%
  Hengkang Wang$^1$ \quad Xu Zhang$^2$\thanks{This work is not related to Dr. Xu Zhang's position at Amazon. } \quad  Taihui Li$^1$ \quad  Yuxiang Wan$^1$ \quad  Tiancong Chen$^1$ \quad  Ju Sun$^1$  \\
  $^1$Department of Computer Science and Engineering, University of Minnesota \\
  \texttt{\{wang9881,lixx5027,wan01530,chen6271,jusun\}@umn.edu} \\
   $^2$Amazon.com Inc., \texttt{spongezhang@gmail.com} \\
}

\begin{document}

\maketitle

\begin{abstract}
    \label{sec:abstract}
Pretrained diffusion models (DMs) have recently been popularly used in solving inverse problems (IPs). The existing methods mostly interleave iterative steps in the reverse diffusion process and iterative steps to bring the iterates closer to satisfying the measurement constraint. However, such interleaving methods struggle to produce final results that look like natural objects of interest (i.e., manifold feasibility) and fit the measurement (i.e., measurement feasibility), especially for nonlinear IPs. Moreover, their capabilities to deal with noisy IPs with unknown types and levels of measurement noise are unknown. In this paper, we advocate viewing the reverse process in DMs as a function and propose a novel plug-in method for solving IPs using pretrained DMs, dubbed DMPlug. DMPlug addresses the issues of manifold feasibility and measurement feasibility in a principled manner, and also shows great potential for being robust to unknown types and levels of noise. Through extensive experiments across various IP tasks, including two linear and three nonlinear IPs, we demonstrate that DMPlug consistently outperforms state-of-the-art methods, often by large margins especially for nonlinear IPs. The code is available at \url{https://github.com/sun-umn/DMPlug}. 
\end{abstract}

\section{Introduction}
\label{sec:intro}
Inverse problems (IPs) are prevalent in numerous fields, such as computer vision, medical imaging, remote sensing, and autonomous driving~\cite{janai_computer_2021,szeliski_computer_2022,olsen_introduction_2007,sylvester_introduction_2005}. The goal of IPs is to recover an unknown object $\mb x$ from noisy measurements $\mb y = \mc{A} \paren{\mb x} + \mb n$, where $\mc{A}$ is a (possibly nonlinear) forward model and $\mb n$ denotes the measurement noise. IPs are often ill-posed: typically, even if $\mb y$ is noiseless, $\mb x$ cannot be uniquely determined from $\mb y$ and $\mc A$. Hence, incorporating prior knowledge on $\mb x$ is necessary to obtain a reliable estimate of the underlying $\mb x$. 

Traditionally, IPs are solved in the regularized data-fitting framework, often motivated as performing the Maximum a Posterior (MAP) inference: 
\begin{align} \label{eq:MAP}
    \min\nolimits_{\mb x}\; \ell\paren{\mb y, \mc{A}\paren{\mb x}} + \Omega\paren{\mb x} \quad \quad \ell\paren{\mb y, \mc{A}\paren{\mb x}}: \text{data-fitting loss}, \;  \Omega\paren{\mb x}: \text{regularizer}.
\end{align}
Here, minimizing the data-fitting loss promotes $\mb y \approx \mc{A}\paren{\mb x}$, and the regularizer encodes prior knowledge on $\mb x$. Recently, the advent of deep learning (DL) has brought about a few new algorithmic ideas to solve IPs. For example, given a training set of measurement-object pairs, i.e., $\{(\mb y_i, \mb x_i)\}_{i=1, \dots, N}$, one can hope to train a DL model that directly predicts $\mb x$ for a given $\mb y$~\cite{kupyn_deblurgan-v2_2019,tsai_stripformer_2022,zamir_multi-stage_2021,delbracio2023inversion,gilton_deep_2021,luo_image_2023,guo_shadowdiffusion_2022,whang_deblurring_2021,gao_implicit_2023,kupyn_deblurgan_2018,ongie_deep_2020,monga_algorithm_2020}. However, such hopes can be shattered by practical challenges in collecting \textbf{massive and realistic} paired training sets, especially for complex IPs~\cite{zhang_deep_2022,koh_single-image_2021}. Even if such challenges can be tackled, one may need to collect a new training set and train a new DL model for every new IP~\cite{ongie_deep_2020}, overlooking potential shared priors on $\mb x$ across IPs. An attractive alternative family of ideas combine pretrained priors on $\mb x$ and regularized data-fitting in \cref{eq:MAP}. For example, they first model the distribution of $\mb x$ using deep generative models, such as generative adversarial networks (GANs) and diffusion models (DMs), based on training sets of the form $\{\mb x_i\}_{i=1, \dots, N}$, and then encode these pretrained generative priors when solving \cref{eq:MAP}. In this way, pretrained priors on $\mb x$ can be reused in an off-the-shelf manner in different IPs about the same family of structured objects.  

\begin{figure}[t]
    \centering 
    \vspace{-1em}
    \includegraphics[width=\linewidth]{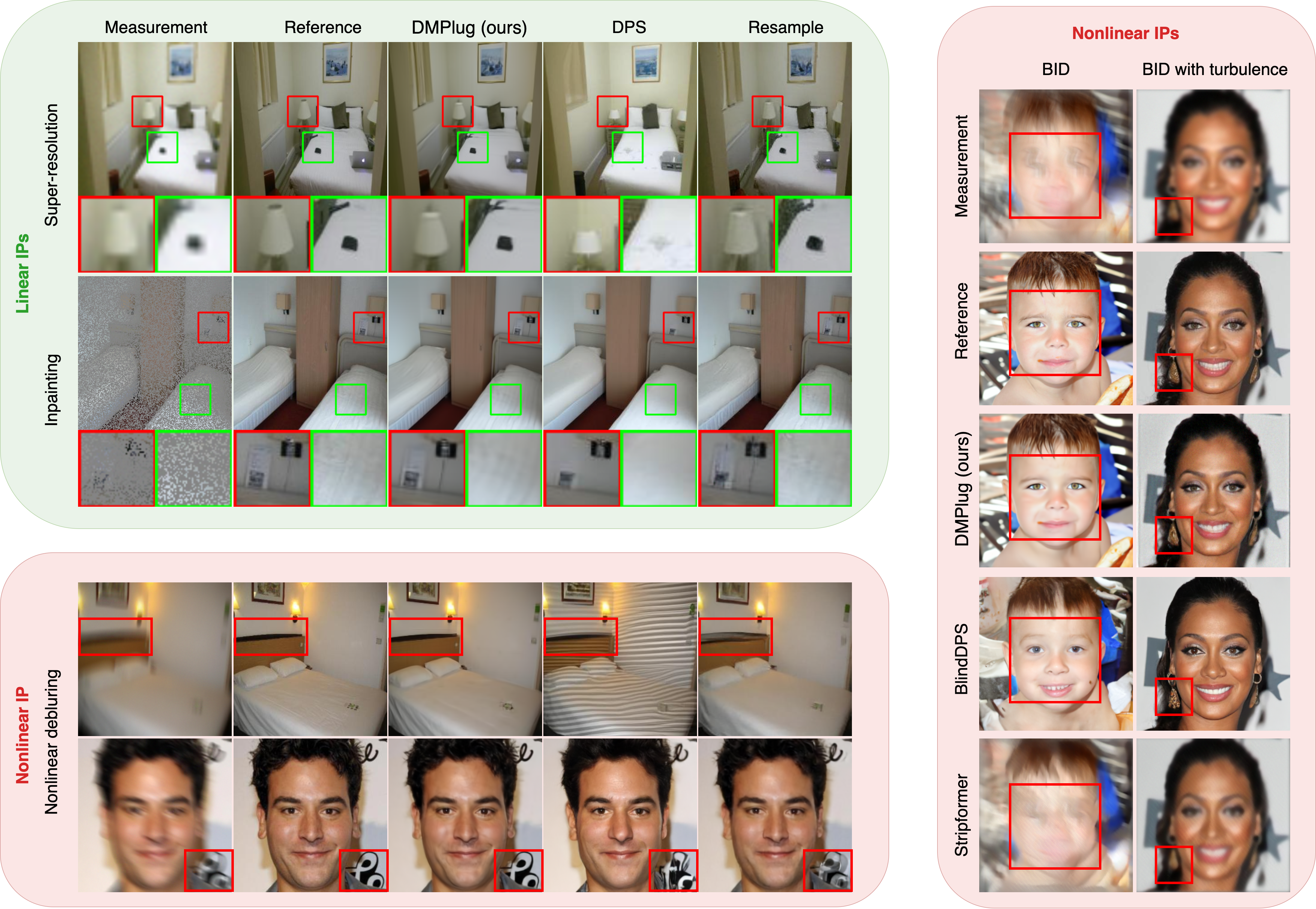}
    \vspace{-1em}
    \caption{Visualization of sample results from our DMPlug method (\textbf{Ours}) and main competing methods (\textbf{DPS}~\cite{chung_diffusion_2023} and \textbf{Resample}~\cite{song_solving_2023} for super-resolution, inpainting, and nonlinear deblurring; \textbf{BlindDPS}~\cite{chung_parallel_2022} and \textbf{Stripformer}~\cite{tsai_stripformer_2022} for blind image deblurring (BID) and BID with turbulence) on IPs we focus on in this paper. All measurements contain Gaussian noise with $\sigma = 0.01$.}
    \label{fig:ips}
\end{figure}

\begin{wrapfigure}{r}{0.45\textwidth}
    \vspace{-1em}
    \centering
    \includegraphics[width=0.95\linewidth]{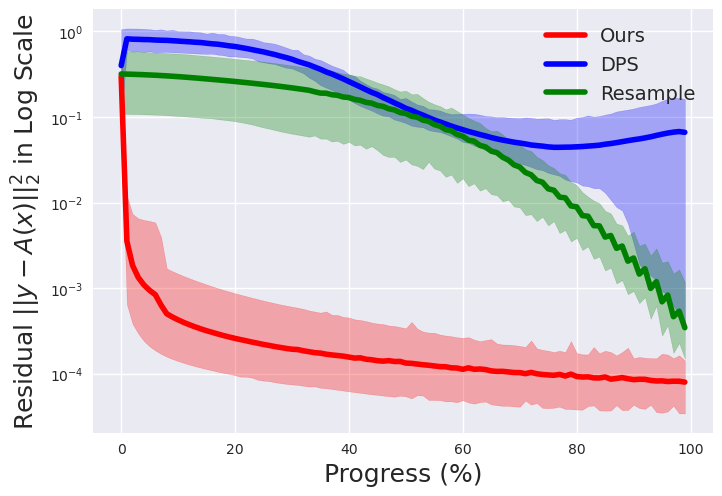}
    \vspace{-1em}
    \caption{\textbf{Evolution of the data-fitting loss $\| \mb y - \mc{A}\paren{\mb x} \|_2^2$ of our DMPlug method vs. SOTA methods over percentage progress}, for noiseless nonlinear deblurring on the CelebA dataset. Here, the percentage progress is calculated with respect to the total number of iterations taken by each method. The shadow regions indicate the ranges of the loss over $50$ instances.}
    \label{fig:issue}
\end{wrapfigure}
\begin{figure}[!htbp]
    \centering
    \vspace{-1em}
    \begin{minipage}{.63\textwidth}
        \centering
        \includegraphics[width=1\linewidth]{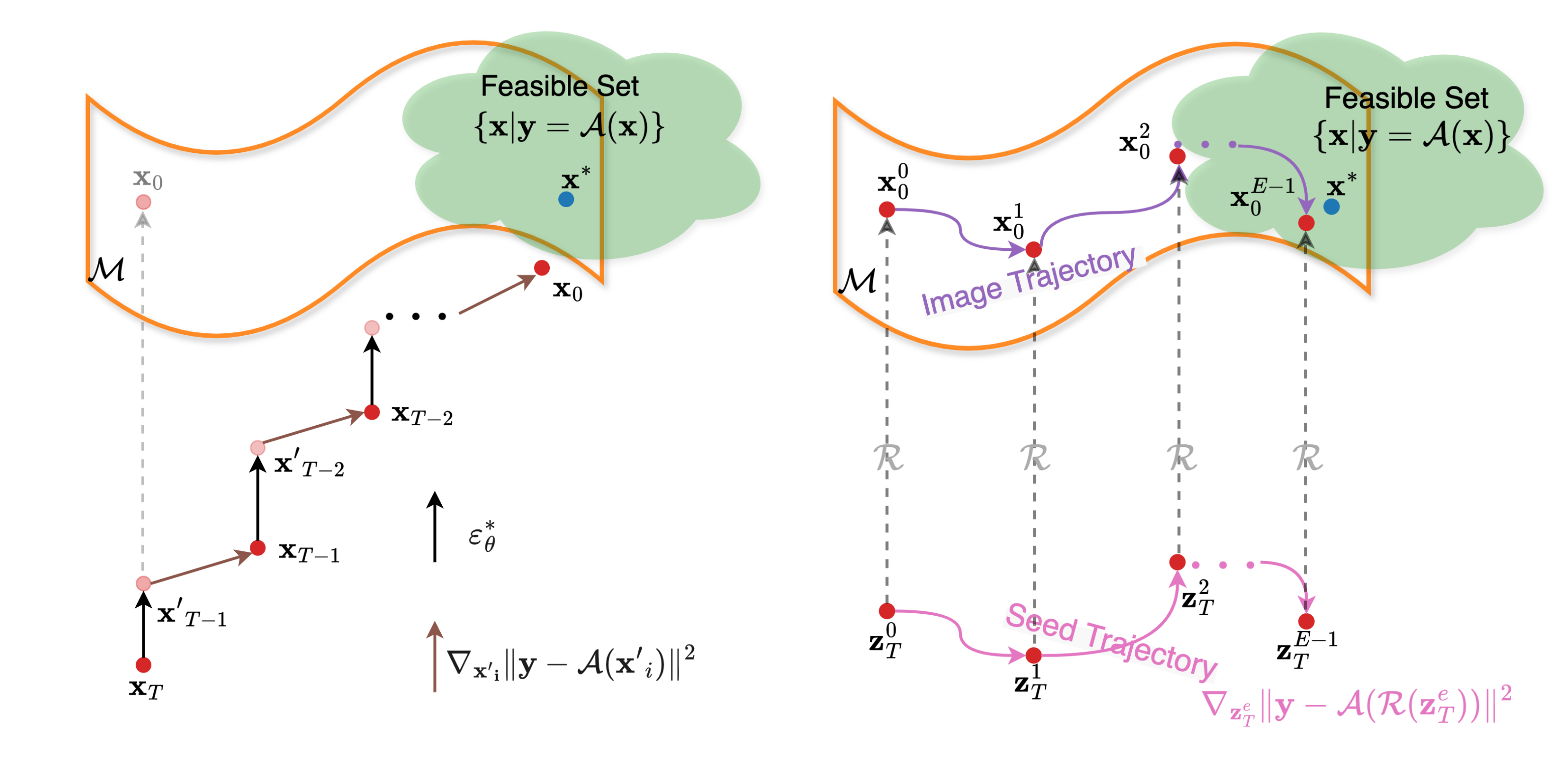}
    \end{minipage}
    \begin{minipage}{.36\textwidth}
        \centering
        \caption{Interleaving methods (left) vs. our DMPlug method (right) for solving IPs using pretrained DMs. While interleaving methods cannot ensure the feasibility of the final estimate for either the object manifold $\mc M$ or the feasible set $\set{\mb x | \mb y = \mc A(\mb x)}$, our DMPlug method ensures the manifold feasibility while promoting $\mb y \approx \mc A(\mb x)$ through global optimization.}
        \label{fig:interleave_plugin} 
    \end{minipage}
    \vspace{-0.5em}
\end{figure}

\textbf{In this paper, we focus on solving IPs with pretrained DMs}. DMs have recently emerged as a dominant family of deep generative models due to their relative stability during training (e.g., vs. GANs) and their strong capabilities to generate photorealistic images (and, depending on applications, other structured objects) once trained~\cite{dhariwal_diffusion_2021,ho_denoising_2020,song_denoising_2022,rombach_high-resolution_2022,he_manifold_2023,xu_provably_nodate}. These strengths of DMs have motivated ideas to use pretrained DMs to solve IPs, such as denoising, super-resolution, inpainting, deblurring, and phase retrieval~\cite{zhu_denoising_2023,meng_diffusion_2024,song_solving_2023,chung_improving_2022,li_diffusion_2023,kawar_denoising_2022,wang_zero-shot_2022,chung_parallel_2022,he_manifold_2023,xu_provably_nodate,liu_accelerating_2024}. Most of these ideas interleave iterative reverse diffusion steps and iterative steps (sometimes projection steps, especially for linear IPs) to move closer to the feasibility set $\{\mb x | \mb y = \mc A \paren{\mb x}\}$ (see \cref{fig:interleave_plugin} and \cref{alg:two_algorithms} (1)). 
However, they typically cannot guarantee the final convergence of the iteration sequence to either the feasible set (i.e., \textbf{measurement feasibility}), or the object manifold $\mc M$ captured by pretrained DMs (i.e., \textbf{manifold feasibility}), as they have modified both iterative processes (see detailed arguments in \cref{sec:bg}). \cref{fig:ips} shows visible artifacts produced by these ideas on several IPs, highlighting \textbf{\textcolor{blue}{(Issue 1)} insufficient manifold feasibility}. To quickly confirm \textbf{\textcolor{blue}{(Issue 2)} insufficient measurement feasibility}, we experiment on \textbf{noiseless} instances of the nonlinear deblurring problem (a nonlinear IP) following~\cite{song_solving_2023,chung_diffusion_2023}, and find that state-of-the-art (SOTA) methods fail to find an $\mb x$ that satisfies $\mb y = \mc{A}\paren{\mb x}$, as shown in \cref{fig:issue}. Furthermore, most SOTA DM-based methods for IPs assume known noise types (e.g., often Gaussian) and known, often very low, noise levels, casting doubts on their performance when faced with unknown noise types and levels, i.e., \textbf{\textcolor{blue}{(Issue 3)} robustness to unknown noise types and levels}, as we confirm in \cref{tab:robust_es}. 

\paragraph{\textcolor{umn_maroon}{Our contributions}} 
In this paper, \textbf{we propose a novel plug-in method, dubbed DMPlug, to solve IPs with pretrained DMs to mitigate all of the above three issues}. DMPlug departs from the popular and dominant interleaving line of ideas by viewing the whole reverse diffusion process as a function $\mc R(\cdot)$, consisting of multiblock stacked DL models, mapping from the seed space to the object manifold $\mc M$. This novel perspective allows us to naturally parameterize the object to be recovered as $\mb x = \mc R\paren{\mb z}$ and then plug this reparametrization into \cref{eq:MAP}, leading to a unified optimization formulation \textbf{with respect to the seed $\mb z$}. Conceptually, since $\mc R(\mb z)$ probably produces a feasible point on the object manifold $\mc M$ and global optimization of the unified formulation encourages $\mb y \approx \mc A(\mc R(\mb z))$, the optimized $\mb z$ could lead to an $\mb x = \mc R(\mb z)$ that \textbf{enjoys both manifold and measurement feasibility}, i.e., \textbf{\textcolor{blue}{(tackling Issues 1 \& 2)}}. \cref{fig:interleave_plugin} and \cref{alg:two_algorithms} schematically illustrate the dramatic difference between interleaving methods and our plug-in method (DMPlug), and \cref{fig:ips,fig:issue} confirms DMPlug's \textbf{strong capability} in finding feasible solutions. For noisy IPs with unknown noise types and levels, we observe that our DmPlug enjoys a benign ``early-learning-then-overfitting'' (ELTO) property:  
the quality of the estimated object climbs first to a peak and then 
degrades once the noise is picked up, as shown in \cref{fig:regression_denoising} (2). This benign property, in combination with appropriate early-stopping (ES) methods that stop the estimate sequence near the peak, allows our method to solve IPs without the exact noise information, i.e., \textbf{\textcolor{blue}{(tackling Issue 3)}}, as shown in \cref{tab:robust_es}). Fortunately, we find that an ES method, ES-WMV~\cite{wang_early_2023} (co-developed by a subset of the current authors), works well for this purpose (see \cref{tab:robust_es}). 

Our contributions can be summarized as follows. \textbf{\textcolor{umn_maroon}{(1)}} In \cref{subsec:formulation}, we \textbf{pioneer a novel plug-in method, DMPlug}, which is significantly different from the prevailing interleaving methods, to solve IPs with pretrained DMs. Then we \textbf{make the proposed method practical} in terms of computational and memory expenses by leveraging one key observation; \textbf{\textcolor{umn_maroon}{(2)}} In \cref{sec:exps} and \cref{app:more_res}, we perform extensive experiments on various linear and nonlinear IPs and show that \textbf{our method outperforms SOTA methods}, both qualitatively and quantitatively---often by large margins, especially on nonlinear IPs. For example, measured in PSNR, our method can lead SOTA methods by about \textcolor{darkpastelgreen}{$2$dB} and \textcolor{darkpastelgreen}{$3 \sim 6$dB} for linear and nonlinear IPs, respectively. Moreover, our method demonstrates flexibility in employing different priors and optimizers, as explored in \cref{subsec:exp_ablation}; \textbf{\textcolor{umn_maroon}{(3)}} In \cref{subsec:issue2}, we \textbf{observe an early-learning-then-overfitting (ELTO) property}, i.e., that our DMPlug tends to recover the desirable object first and then overfit to the potential noise. By taking advantage of this benign property and integrating the ES method ES-WMV~\cite{wang_early_2023}, our method is \textbf{the first to achieve robustness to unknown noise types and levels}, leading SOTA methods by about \textcolor{darkpastelgreen}{$1$dB} and \textcolor{darkpastelgreen}{$3.5$dB} in terms of PSNR for linear and nonlinear IPs, respectively.

\section{Background and related work}
\label{sec:bg}

\paragraph{\textcolor{umn_maroon}{Diffusion models (DMs)}} 
The denoising diffusion probabilistic model (\textbf{DDPM})~\cite{ho_denoising_2020} is a seminal DM for unconditional image generation. It gradually transforms $\mb x_0 \sim p_{\text{data}}$ into total noise $\mb x_T \sim \mc N\paren{\mb 0, \mb I}$ (i.e., forward diffusion process) and then learns to gradually recover $\mb x_0$ from $\mb x_T$ through incremental denoising (i.e., reverse diffusion process). The forward process can be described by a stochastic differential equation (SDE), $d\mb x = -\beta_t/2 \cdot \mb x dt + \sqrt{\beta_t} d\mb w$, where $\beta_t$ is the noise schedule and $\mb w$ is the standard Wiener process. The corresponding reverse process is described by 
\begin{align} \label{eq:reverse_SDE}
(\text{Reverse SDE for DDPM}) \; d\mb x = -\beta_t \left[\mb x/2 + \nabla_{\mb x} \log p_t(\mb x) \right] dt + \sqrt{\beta_t} d\ol{\mb w}.
\end{align} 
Here, $\ol{\mb w}$ is the time-reversed standard Wiener process, $p_t(\mb x)$ is the probability density at time $t$, and $\nabla_{\mb x} \log p_t(\mb x)$ is the (Stein) score function, which is approximated by a DL model {\small $\mb \varepsilon_{\mb \theta}^{(t)}\paren{\mb x}$} via score matching methods during DM training~\cite{hyvarinen_estimation_2005,song_generative_2020}. For discrete settings, given time steps $t \in \{1, \dots, T\}$, a variance schedule $\beta_1, \dots, \beta_T$, $\alpha_t \doteq 1 - \beta_t$ with $\alpha_T \to 0$, and $\bar{\alpha}_t \doteq \prod_{s=1}^t \alpha_s$, the DDPM has the forward process $\mb x_t = \sqrt{1-\beta_t} \mb x_{t-1} + \sqrt{\beta_t} \mb z$ and the reverse process 
\begin{align}\label{eq:ddpm}
   (\text{DDPM}) \;  \mb x_{t-1} & = 1/\sqrt{\alpha_t} \cdot \left(\mb x_t - \beta_t/\sqrt{1-\bar{\alpha}_t} \cdot \mb \varepsilon_{\mb \theta}^{(t)}(\mb x_t) \right) + \sqrt{\beta_t} \mb z,  
\end{align}
where $\mb z \sim \mc N\paren{\mb 0, \mb I}$. As the DDPM 
has a slow reverse/sampling process, \cite{song_denoising_2022} proposes the denoising diffusion implicit model (\textbf{DDIM}) to mitigate this issue. With the same notation as that of the DDPM, the DDIM makes a crucial change to the DDPM: relaxing the forward process to be non-Markovian by making $\mb x_t$ depend on both $\mb x_0$ and $\mb x_{t-1}$. This simple change allows skipping iterative steps in the reverse process, without retraining the DDPM. This leads to much smaller numbers of reverse steps, and hence substantial speedup in sampling. The reverse process is now defined as
\begin{align}\label{eq:ddim}
    \hspace{-1em} (\text{DDIM}) \; \mb x_{t-1} \hspace{-0.2em} = \hspace{-0.2em}\sqrt{\bar{\alpha}_{t-1}} \widehat{\mb x}_0\paren{\mb x_t}
   \hspace{-0.2em} + \hspace{-0.2em} {\sqrt{1-\bar{\alpha}_{t-1}}} \mb \varepsilon_{\mb \theta}^{(t)}(\mb x_t), 
\end{align}
where $\widehat{\mb x}_0\paren{\mb x_t} \doteq [\mb x_t - \sqrt{1-\bar{\alpha}_t} \mb \varepsilon_{\mb \theta}^{(t)}(\mb x_t)]/\sqrt{\bar{\alpha}_t}$ is the predicted $\mb x_0$ with $\mb x_t$. 

\paragraph{\textcolor{umn_maroon}{Pretrained DMs for solving IPs}} 
Ideas for solving IPs with DMs can be classified into two categories: supervised and zero-shot~\cite{li_diffusion_2023,song_DMtutorial_2023}. The former trains DM-based IP solvers based on paired training sets of the form $\{(\mb y_i, \mb x_i)\}$ and is hence not our focus here (see our arguments in \cref{sec:intro}). The latter makes use of pretrained DMs as data-driven priors: \textbf{\textcolor{umn_maroon}{(I)}} Most of the work in this category considers modeling $p_t(\mb x| \mb y)$ directly and replaces the (unconditional) score function $\nabla_{\mb x} \log p_t(\mb x)$ in \cref{eq:reverse_SDE} by the conditional score function $\nabla_{\mb x} \log p_t(\mb x | \mb y) =  \nabla_{\mb x} \log p_t(\mb x) + \nabla_{\mb x} \log p_t(\mb y | \mb x)$, leading to the conditional reverse SDE
\begin{align} \label{eq:cond_reverse_sde}
d\mb x = \left[ -\beta_t/2 \cdot\mb x - \beta_t \paren{\nabla_{\mb x} \log p_t(\mb x) + \nabla_{\mb x} \log p_t(\mb y | \mb x)} \right] dt + \sqrt{\beta_t} d \ol{\mb w}.
\end{align}
Here, while $\nabla_{\mb x} \log p_t(\mb x)$ can be naturally approximated by the pretrained score function $\mb \varepsilon_{\mb \theta}^{(t)}\paren{\mb x}$, $\nabla_{\mb x} \log p_t(\mb y | \mb x)$ is intractable as $\mb y$ does not directly depend on $\mb x(t)$\footnote{Recall that in the continuous formulation of diffusion processes, $\mb x$ is a function of $t$, i.e., $\mb x(t)$ where the continuous variable $t \in [0, T]$ is often omitted. }. Ideas to circumvent this difficulty include approximating $p_t(\mb y | \mb x(t))$ by $p_t(\mb y | \widehat{\mb x}(0)[\mb x(t)])$, where $\widehat{\mb x}(0)[\mb x(t)]$ is implemented as $\widehat{\mb x}_0\paren{\mb x_t}$ of \cref{eq:ddim} in discretization~\cite{chung_diffusion_2023,chung_parallel_2022}, and interleaving unconditional reverse steps of \cref{eq:ddpm} or \cref{eq:ddim} and (approximate) projections onto the feasible set $\{\mb x | \mb y = \mc A \paren{\mb x}\}$ to bypass the likelihood $p_t(\mb y | \mb x)$~\cite{kadkhodaie_stochastic_2021,chung_improving_2022,wang_zero-shot_2022,kawar_denoising_2022,song_score-based_2021,choi_ilvr_2021,song_solving_2023}; \textbf{\textcolor{umn_maroon}{(II)}} An interesting alternative is to recall that the MAP framework (see also \cref{eq:MAP}) involves $\max_{\mb x}\; \log p(\mb y | \mb x) + \log p(\mb x)$, and first-order methods, especially proximal-gradient style methods, to optimize the MAP formulation typically only need to access $p(\mb x)$ through $\nabla_{\mb x} \log p(\mb x)$, i.e., the score function---the central object in DMs! So, one can derive IP solvers by wrapping pretrained DMs around first-order methods for (approximately) optimizing the MAP formulation, see, e.g., ~\cite{xu_provably_nodate,zhu_denoising_2023}. 
\begin{figure}[!htbp]
\vspace{-1em}
\centering
\begin{minipage}{0.47\textwidth}
\begin{algorithm}[H]
\caption{Template for interleaving methods}
\label{alg:interleaving}
\begin{algorithmic}[1]
\Require \# Diffusion steps $T$, measurement $\mb y$
\State $\mb x_T \sim \mc N\paren{\mb 0, \mb I}$
\For{$i = T - 1$ to $0$}
    \State $\hat{\mb s} \gets \mb \varepsilon_{\mb \theta}^{(i)}\paren{\mb x_i}$
    \State $\hat{\mb x}_0 \gets \frac{1}{\sqrt{\Bar{\alpha}_i}}\paren{\mb x_i - \sqrt{1 - \Bar{\alpha}_i} \hat{\mb s}}$
    \State $\mb x'_{i-1} \gets$ DDIM reverse with $\hat{\mb x}_0$ and $\hat{\mb s}$
    \State $\mb x_{i-1} \gets$ \textcolor{blue}{(Approximately) Projection}~\cite{kadkhodaie_stochastic_2021,chung_improving_2022,wang_zero-shot_2022,kawar_denoising_2022,song_score-based_2021,choi_ilvr_2021,liu_accelerating_2024} or \textcolor{blue}{gradient update}~\cite{song_solving_2023,zhu_denoising_2023,chung_diffusion_2023,chung_parallel_2022,meng_diffusion_2024,xu_provably_nodate,he_manifold_2023} with $\hat{\mb x}_0$ and $\mb x'_{i-1}$ to get closer to $\set{\mb x | \mb y = \mc A(\mb x)}$
\EndFor
\Ensure Recovered object $\mb x_0$
\end{algorithmic}
\end{algorithm}
\end{minipage}
\begin{minipage}{0.52\textwidth}
\begin{algorithm}[H]
\caption{Proposed plug-in method, DMPlug}
\begin{algorithmic}[1]
\Require \# Epochs $E$, \# diffusion steps $T$, $\mb y$
\State Initialize seed $\mb z_T^0 \sim \mc N\paren{\mb 0, \mb I}$
\For{$e = 0$ to $E - 1$}
    \For{$i = T - 1$ to $0$} \tikzmark{top}
    \State $\hat{\mb s} \gets \mb \varepsilon_{\mb \theta}^{(i)}\paren{\mb z_i^e}$
    \State $\hat{\mb z}_0^e \gets \frac{1}{\sqrt{\Bar{\alpha}_i}}\paren{\mb z_i^e - \sqrt{1 - \Bar{\alpha}_i} \hat{\mb s}}$
    \State $\mb z_{i-1}^e \gets$ DDIM reverse with $\hat{\mb z}_0^e, \hat{\mb s}$ \tikzmark{right}
    \EndFor \tikzmark{bottom} 
    \State Update $\mb z_T^{e+1}$ from $\mb z_T^{e}$ via a gradient update for \cref{eq:ours}
\EndFor
\Ensure Recovered object $\mc R\paren{\mb z_T^{E-1}}$
\AddNote{top}{bottom}{right}{$\; \mc R$}
\end{algorithmic}
\end{algorithm}
\end{minipage}
\caption{Comparison of prevailing interleaving methods and our plug-in method}
\label{alg:two_algorithms}
\vspace{-1em}
\end{figure} 

\textbf{Despite the disparate conceptual ways of utilizing pretrained DM priors, most of the methods under \textbf{(I) and (II)} proposed in the literature so far follow a single algorithmic template, i.e., \cref{alg:interleaving}}. There, Lines 3-5 are simply a reverse iterative step in the DDIM (\cref{eq:ddim}; obviously, one could replace this by a reverse iterative step in other appropriate pretrained DMs.  Line 6 helps move the iterate closer or onto the feasible set $\{\mb x | \mb y = \mc A \paren{\mb x}\}$. In other words, these methods interleave iterative steps to move toward the data manifold $\mc M$ defined by the pretrained DM and iterative steps to move toward the feasible set $\{\mb x | \mb y = \mc A \paren{\mb x}\}$, i.e., as illustrated in \cref{fig:interleave_plugin} (left). However, it is unclear, a priori, whether such interleaving iterative sequence will converge to either, leading to concerns about \textbf{\textcolor{blue}{(Issue 1)} insufficient manifold feasibility} and \textbf{\textcolor{blue}{(Issue 2)} insufficient measurement feasibility}. In fact, our \cref{fig:ips,fig:issue} confirm these issues empirically, echoing observations made in several prior papers~\cite{song_solving_2023,sanghvi2023kernel,chung_improving_2022}. In principle, Issue 2 can be mitigated by ensuring that Line 6 finds a feasible $\mb x_{i-1}$ for $\{\mb x | \mb y = \mc A \paren{\mb x}\}$ in each iteration~\cite{kadkhodaie_stochastic_2021,chung_improving_2022,wang_zero-shot_2022,kawar_denoising_2022,song_score-based_2021,choi_ilvr_2021,song_solving_2023}. However, this is possible only for easy IPs (e.g., linear IPs where we can perform a closed-form projection) and is difficult for typical nonlinear IPs as hard nonconvex problems are entailed. Moreover, although most of these methods have considered noisy IPs alongside noiseless ones~\cite{kadkhodaie_stochastic_2021,chung_improving_2022,wang_zero-shot_2022,kawar_denoising_2022,song_solving_2023,zhu_denoising_2023,chung_diffusion_2023,chung_parallel_2022,meng_diffusion_2024}, their common assumption about known noise types (often Gaussian) and levels (often low) is unrealistic: there are many types of measurement noise in practice---often hybrid from multiple sources~\cite{hendrycks_benchmarking_2019}, and the noise levels are usually unknown and hard to estimate~\cite{li_single_2023}, raising concerns about  \textbf{\textcolor{blue}{(Issue 3)} the robustness of these methods to unknown noise types and levels}; see \cref{subsec:exp_robust}.

\section{Method}
\label{sec:method}
In this section, we propose a simple plug-in method, DMPlug, to solve IPs with pretrained DMs to address Issues 1 \& 2 in \cref{subsec:formulation}, discuss its connection to and difference from several familiar ideas in \cref{subsec:similar}, and finally explain how to integrate an early-stopping strategy, ES-WMV~\cite{wang_early_2023}, into our plug-in method to address Issue 3 in \cref{subsec:issue2}, leading to an algorithm, DMPlug+ES-WMV, summarized in \cref{alg:es}, that solves IPs with pretrained DMs under the regularized data-fitting framework~\cref{eq:MAP}, even in the presence of unknown noise. 

\subsection{Our plug-in method: DMPlug}
\label{subsec:formulation}
\paragraph{\textcolor{umn_maroon}{Reverse process as a function and our plug-in method}}  
Interleaving methods discussed above can have trouble satisfying both manifold feasibility and measurement feasibility because they interleave and hence modify both processes. Although projection-style modifications can improve measurement feasibility~\cite{kadkhodaie_stochastic_2021,chung_improving_2022,wang_zero-shot_2022,kawar_denoising_2022,song_score-based_2021,choi_ilvr_2021,song_solving_2023}, their application to nonlinear IPs seems tricky, and a single ``projection'' step could be as difficult and expensive as solving the original problem due to the typical nonconvexity induced by the nonlinearity in $\mc A$~\cite{zhu_denoising_2023,song_solving_2023}. In contrast, we propose \textbf{viewing the whole reverse process as a function $\mc R\paren{\cdot}$} that maps from the seed space to the object space (or the object manifold $\mc M$). Mathematically, if we write a single inverse step as a function $g$ that depends on {\small $\mb \varepsilon_{\mb \theta}^{(i)}$}, i.e., {\small $g_{\mb \varepsilon_{\mb \theta}^{(i)}}$} for the $i$-th reverse step that maps $\mb x_{i+1}$ to $\mb x_{i}$, then 
\begin{align} \label{eq:R_def}
    \mc R = g_{\mb \varepsilon_{\mb \theta}^{(0)}} \circ g_{\mb \varepsilon_{\mb \theta}^{(1)}} \circ \cdots \circ g_{\mb \varepsilon_{\mb \theta}^{(T-2)}} \circ g_{\mb \varepsilon_{\mb \theta}^{(T-1)}}.   \quad \quad (\text{$\circ$ means function composition})
\end{align}
Note that for the DDPM, and those DMs based on SDEs in general, $\mc R$ is a stochastic function due to noise injection in each step. To reduce technicality, we focus on DMs based on ordinary differential equations (ODEs), in particular, the DDIM, due to their increasing popularity~\cite{lu_dpm-solver_2022,zheng_dpm-solver-v3_2023}, resulting in deterministic $\mc R$'s. This conceptual leap allows us to reparametrize our object of interest as $\mb x = \mc R(\mb z)$ and plug this reparametrization into the traditional regularized data-fitting framework in \cref{eq:MAP}, yielding the following unified optimization formulation:  
\begin{align} \label{eq:ours}
    (\text{\textbf{DMPlug}}) \; \mb z^* \in \argmin\nolimits_{\mb z}\; \ell\paren{\mb y, \mc{A}\paren{\mc R\paren{\mb z}}} + \Omega\paren{\mc R\paren{\mb z}}, \quad \quad \mb x^* = \mc{R}\paren{\mb z^*}. 
\end{align}
\textbf{We stress that here the optimization is with respect to the seed variable $\mb z$, given the pretrained reverse process $\mc R$}. Since we never modify the reverse diffusion process $\mc R$, we expect $\mc R(\mb z)$ to produce an object on the object manifold $\mc M$, i.e., enforcing manifold feasibility to \textbf{\textcolor{blue}{address Issue 1}}. Moreover, optimizing the unified formulation \cref{eq:ours} is expected to promote $\mb y \approx \mc{A}\paren{\mc R(\mb z^*)}$, inherent in the regularized data-fitting framework, i.e., promoting data feasibility to \textbf{\textcolor{blue}{address Issue 2}}. 

When there are multiple objects of interest in the IP under consideration, i.e.,  $\mb y = \mc A(\mb x_1, \cdots, \mb x_k) + \mb n$ with $k$ objects (e.g., in blind image deblurring $\mb y = \mc A(\mb k , \mb x) + \mb n$, both the blur kernel $\mb k$ and sharp image $\mb x$ are objects of interest, where $\mc A(\mb k, \mb x) = \mb k \ast \mb x$; see also \cref{app:bid}), different objects may have different priors that should be encoded and treated differently. Our unified optimization formulation \cref{eq:ours} facilitates the natural integration of multiple priors. Another bonus feature of our unified formulation lies in the flexibility in choosing numerical optimization solvers. We briefly explore both aspects in \cref{subsec:exp_ablation}. 

\begin{wrapfigure}{r}{0.35\textwidth}
    \vspace{-1em}
    \centering
    \includegraphics[width=1\linewidth]{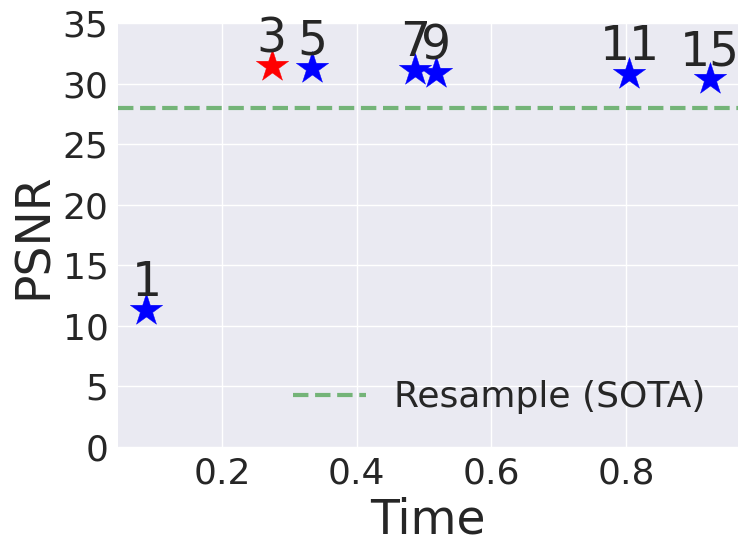}
    \vspace{-2em}
    \caption{\footnotesize{\textbf{PSNR (dB) vs. per-iteration wall-clock time (s)} running on an \textit{NVIDIA A100}, for various reverse steps in $\mc{R}$. Experiments on CelebA for $4 \times$ super-resolution; solver: \textit{ADAM}; maximum iterations: $6,000$}}
    \vspace{-1em}
    \label{fig:psnrvstime1}
\end{wrapfigure}

\paragraph{\textcolor{umn_maroon}{Fast samplers for memory and computational efficiency}} 
When implementing DMPlug with typical gradient-based solvers, e.g., \textit{ADAM}~\cite{kingma_adam_2017}, the gradient calculation in each iterative step requires a forward and a backward pass through the entire $\mc R$, i.e., $T$ blocks of the basic DL model in the given DM. For high-quality image generation~\cite{ho_denoising_2020,song_denoising_2022,rombach_high-resolution_2022}, $T$ is typically tens or hundreds, resulting in prohibitive memory and computational burdens. To resolve this, we use the DDIM, which allows skipping reverse sampling steps thanks to its non-Markovian property, as the sampler $\mc{R}$. We observe, \emph{to our surprise and also in our favor}, that a very small number of reverse steps, such as $3$, is sufficient for our method to beat SOTA methods on all IPs we evaluate (see \cref{fig:psnrvstime1} and \cref{sec:exps}), and further increasing the number does not substantially improve the performance (actually even slightly degrades it perhaps due to the numerical difficulty caused by vanishing gradients as more steps are included). So, \textbf{we default the number of reverse steps to $3$ unless otherwise stated}. This number is not sufficient for generating high-quality photorealistic images with existing DMs, but is good enough for our method. The discrepancy suggests a fundamental difference between image generation and image ``regression'' involved in solving IPs---we leave this for future work. 

\subsection{Seemingly similar ideas}\label{subsec:similar}

\textbf{\textcolor{umn_maroon}{(A) GAN inversion for IPs}}
Our plug-in method is reminiscent of GAN inversion to solve IPs~\cite{creswell_inverting_2016,zhu_generative_2018}: for a pretrained GAN generator $G_{\mb \theta}$, GAN inversion performs a similar reparametrization $\mb x = G_{\mb \theta} (\mb z)$ and plugs it into \cref{eq:MAP}. The remaining task is also to minimize the resulting formulation with respect to the trainable seed $\mb z$, and produce an $\mb x = G_{\mb \theta}(\mb z^*)$ through a solution $\mb z^*$. However, there are numerous signs that GAN inversion does not work well for IPs. For example, \cite{pan_exploiting_2020} also finetunes the generator $G_{\mb \theta}$ alongside the trainable seed to boost performance, and \cite{wang_zero-shot_2022,kawar_denoising_2022} report superior performance of DM-based methods compared to GAN-inversion-based ones for solving IPs; 
\textbf{\textcolor{umn_maroon}{(B) Diffusion inversion (DI)}} 
Given an object $\mb x$, DI aims to find a seed $\mb z$ so that $\mc R(\mb z)$ reproduces $\mb x$, an important algorithmic component for DM-based image and text editing~\cite{su_dual_2023,hertz_prompt--prompt_2022,kim_diffusionclip_2022,song_denoising_2022,mokady_null-text_2022,wallace_edict_2023}. A popular choice for DI is to modify DDIM~\cite{kim_diffusionclip_2022,song_denoising_2022,mokady_null-text_2022,wallace_edict_2023}. Although we cannot use DI to solve general IPs, DI can be considered as an IP and solved through $\min_{\mb z} \; \ell(\mb x, \mc R(\mb z))$ using our plug-in method. 
\textbf{\textcolor{umn_maroon}{(C) Algorithm unrolling (AU)}} The multiblock structure in $\mc R$ also resembles those DL models used in AU, a popular family of supervised methods for solving IPs~\cite{gregor_learning_2010,monga_algorithm_2020}. In AU, the trainable DL model also consists of multiple blocks of basic DL models, induced by unfolded iterative steps to solve \cref{eq:MAP}. While in AU the weights in these DL blocks are trainable, the weights in our DL blocks are fixed. More importantly, as a supervised approach, AU requires paired training sets of the form $\{(\mb y_i, \mb x_i)\}_{i=1, \dots, N}$, in contrast to the zero-shot nature of our method here.  

\subsection{Achieving robustness to unknown noise}
\label{subsec:issue2}
\begin{figure*}[!htbp]
    \centering 
    \vspace{-1em}
    \includegraphics[width=0.95\linewidth]{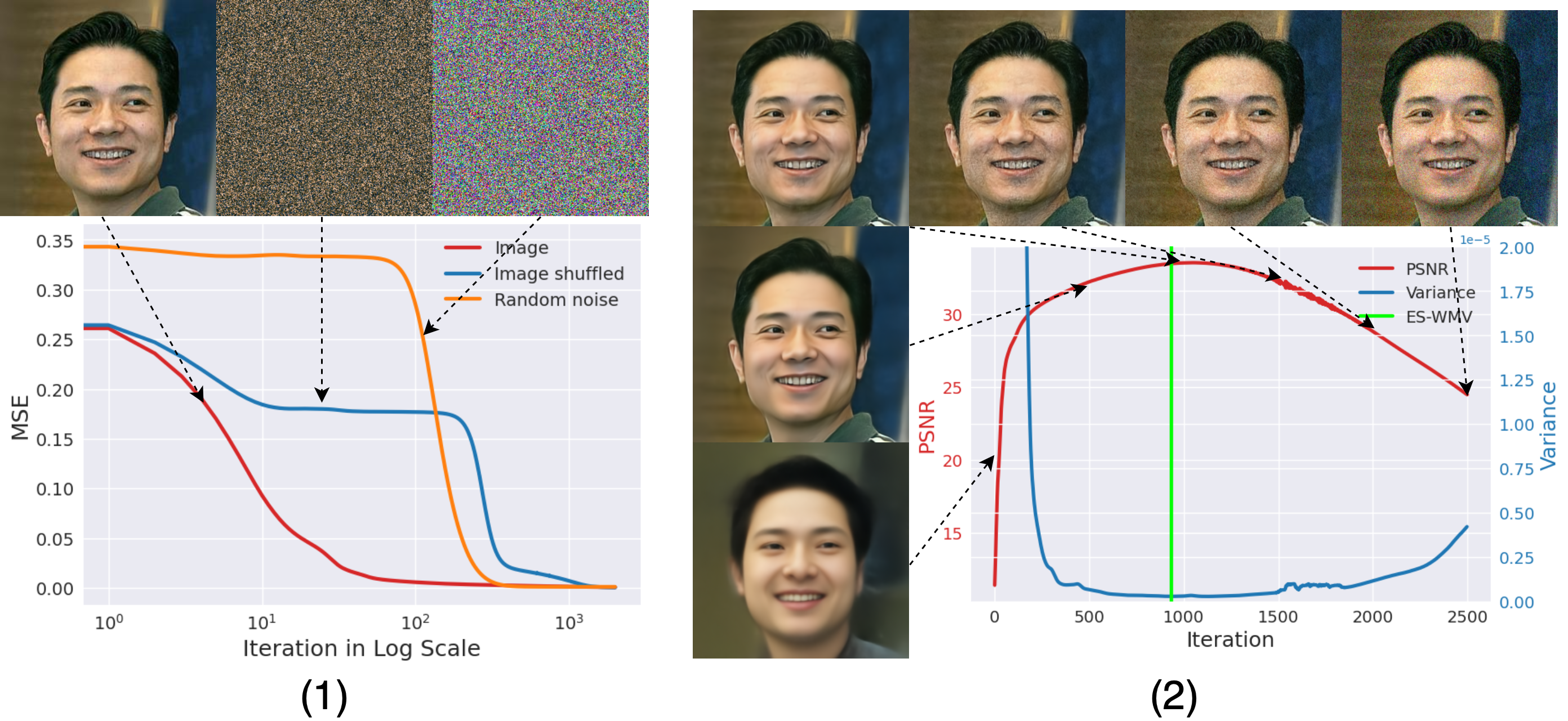}
    \vspace{-1em}
    \caption{The recovery performance of our method on image regression with (1): (a) a clean natural image, (b) the same image with pixels randomly shuffled, (c) random noise iid sampled from $\mathrm{Uniform}\paren{0, 1}$, and (2) a noisy natural image with Gaussian noise at $\sigma = 0.08$. Here, image regression means, given any image $\mb x$, performing $\min_{\mb z}\; \ell\paren{\mb x, \mc R\paren{\mb z}}$.}
    \label{fig:regression_denoising}
\end{figure*}
\paragraph{\textcolor{umn_maroon}{The early-learning-then-overfitting (ELTO) phenomenon}} 
When $\mb y$ contains noise, solving \cref{eq:ours} can promote measurement feasibility, i.e., $\mb y \approx \mc{A}\paren{\mc{R}\paren{\mb z^*}}$, but $\mc{A}\paren{\mc{R}\paren{\mb z^*}}$ may also learn the noise, i.e., \textbf{overfitting to noise}. Interestingly, \textbf{our method seems to favor desirable content and resist noise}: \textbf{\textcolor{umn_maroon}{(A)}} our method converges much faster when used to regress clean natural images than random noise (\cref{fig:regression_denoising} (1), suggesting that our method shows high resistance to noise and low resistance to structured content, and \textbf{\textcolor{umn_maroon}{(B)}} when performing regression against a noisy image $\mb x = \mb x_0 + \mb n$, our method, although powerful enough to overfit the noisy image $\mb x$ ultimately, picks up the desired image content first and then learns the noise, leading to a hallmark ``early-learning-then-overfitting'' (\textbf{ELTO}) phenomenon so that \textbf{the recovery quality climbs to a peak before the potential degradation due to noise} (\cref{fig:regression_denoising} (2)). We stress that similar ELTO phenomena have been widely reported in the literature on using deep image priors (DIPs) to solve IPs~\cite{ulyanov_deep_2020,li_self-validation_2021,li_deep_2023,shi_measuring_2021,zhuang_blind_2024}, although \textbf{this is the first time this phenomenon has been reported for DM-based methods}. Inspired by related studies in DIP, we also perform a spectral analysis of intermediate recovery and reveal that our method has \textbf{a spectral bias toward low-frequency components during learning}, similar to DIP methods; see \cref{app:sb}. 

\paragraph{\textcolor{umn_maroon}{Achieving robustness via early stopping (ES)}} 
One may wonder how widely this ELTO phenomenon occurs when using our DMPlug to solve IPs. Besides extensive additional visual confirmation (see \cref{app:es_res}), in \cref{tab:robust_es} we show that our method, without using the noise information, leads the SOTA methods in terms of peak performance during iteration---particularly, by large margins on non-linear deblurring. The quantitative results suggest that the ELTO phenomenon is likely widespread, especially across noise types and levels. Hence, if we can perform proper early stopping (ES) to locate the peak performance, we \textbf{\textcolor{blue}{tackle Issue 3}}. Deriving an effective ES strategy here is nontrivial, as in practice we do not have groundtruth images to compute any reference-based performance metrics such as PSNR. Fortunately, in the DIP literature, \cite{wang_early_2023} discovers that for DIP-based methods for IPs, the valleys of the running-variance curves of the intermediate reconstructions are well aligned with the performance peaks. Based on this crucial observation, they propose an ES strategy, ES-WMV, that can accurately detect performance peaks with small performance loss on various IP tasks. Inspired by their success, we integrate the ES-WMV strategy into our plug-in method and find that ES-WMV is highly synergetic with our method and performs reliable ES with negligible performance loss (see \cref{tab:robust_es}). Details of the entire algorithm can be found in \cref{alg:es}.

\section{Experiments}
\label{sec:exps}
In this section, we evaluate our plug-in method, DMPlug, and compare it with other SOTA methods on two linear IPs, including \textbf{super-resolution} and \textbf{inpainting}, and three nonlinear IPs, including \textbf{nonlinear deblurring}, \textbf{blind image deblurring (BID)}, and \textbf{BID with turbulence}. Following~\cite{song_solving_2023}, we construct the evaluation sets by sampling $100$ images from CelebA~\cite{liu_deep_2015}, FFHQ~\cite{karras_style-based_2019} and LSUN-bedroom~\cite{yu_lsun_2016}, respectively, and resizing all images to $256 \times 256$; we measure recovery quality using three standard metrics for image restoration, including peak signal-to-noise-ratio (PSNR), structural similarity index (SSIM) and Learned Perceptual Image Patch Similarity (LPIPS)~\cite{zhang_unreasonable_2018} with the default backbone. We describe in detail the five IPs tested and the formulations we use for them in \cref{app:setup}; we provide implementation details of our method and the competing methods in \cref{app:imp}; we include comparisons of computational costs, along with more quantitative and qualitative results in \cref{app:more_res}.  

\vspace{-1em}
\begin{table}[!htpb]
\caption{(\textcolor{red}{Nonlinear IP}) \textbf{Nonlinear deblurring} with additive Gaussian noise ($\sigma = 0.01$). (\textbf{Bold}: best, \underbar{under}: second best, \textcolor{darkpastelgreen}{green}: performance increase, \textcolor{red}{red}: performance decrease)}
\vspace{-1em}
\label{tab:nonuniform}
\begin{center}
\setlength{\tabcolsep}{1.0mm}{
\begin{tabular}{c c c c c c c c c c}
\hline

&\multicolumn{3}{c}{\scriptsize{\textbf{CelebA~\cite{liu_deep_2015} ($256 \times 256$)}}}
&\multicolumn{3}{c}{\scriptsize{\textbf{FFHQ~\cite{karras_style-based_2019} ($256 \times 256$)}}}
&\multicolumn{3}{c}{\scriptsize{\textbf{LSUN~\cite{yu_lsun_2016} ($256 \times 256$)}}}
\\

\cmidrule(lr){2-4}
\cmidrule(lr){5-7}
\cmidrule(lr){8-10}

&\scriptsize{LPIPS$\downarrow$}
&\scriptsize{PSNR$\uparrow$}
&\scriptsize{SSIM$\uparrow$}

&\scriptsize{LPIPS$\downarrow$}
&\scriptsize{PSNR$\uparrow$}
&\scriptsize{SSIM$\uparrow$}

&\scriptsize{LPIPS$\downarrow$}
&\scriptsize{PSNR$\uparrow$}
&\scriptsize{SSIM$\uparrow$}

\\
\hline

\scriptsize{BKS-styleGAN~\cite{tran_explore_2021}}
&\scriptsize{1.047}
&\scriptsize{22.82}
&\scriptsize{0.653}

&\scriptsize{1.051}
&\scriptsize{22.07}
&\scriptsize{0.620}

&\scriptsize{0.987}
&\scriptsize{20.90}
&\scriptsize{0.538}
\\

\scriptsize{BKS-generic~\cite{tran_explore_2021}}
&\scriptsize{1.051}
&\scriptsize{21.04}
&\scriptsize{0.591}

&\scriptsize{1.056}
&\scriptsize{20.76}
&\scriptsize{0.583}

&\scriptsize{0.994}
&\scriptsize{18.55}
&\scriptsize{0.481}
\\

\scriptsize{MCG~\cite{chung_improving_2022}}
&\scriptsize{0.705}
&\scriptsize{13.18}
&\scriptsize{0.135}

&\scriptsize{0.675}
&\scriptsize{13.71}
&\scriptsize{0.167}

&\scriptsize{0.698}
&\scriptsize{14.28}
&\scriptsize{0.188}
\\

\scriptsize{ILVR~\cite{choi_ilvr_2021}}
&\scriptsize{0.335}
&\scriptsize{21.08}
&\scriptsize{0.586}

&\scriptsize{0.374}
&\scriptsize{20.40}
&\scriptsize{0.556}

&\scriptsize{0.482}
&\scriptsize{18.76}
&\scriptsize{0.444}

\\

\scriptsize{DPS~\cite{chung_diffusion_2023}} 
&\scriptsize{0.149}
&\scriptsize{24.57}
&\scriptsize{0.723}

&\scriptsize{0.130}
&\scriptsize{25.00}
&\scriptsize{0.759}

&\scriptsize{0.244}
&\scriptsize{23.46}
&\scriptsize{0.684}
\\

\scriptsize{ReSample~\cite{song_solving_2023}} 
&\scriptsize{\underbar{0.104}}
&\scriptsize{\underbar{28.52}}
&\scriptsize{\underbar{0.839}}

&\scriptsize{\underbar{0.104}}
&\scriptsize{\underbar{27.02}}
&\scriptsize{\underbar{0.834}}

&\scriptsize{\underbar{0.143}}
&\scriptsize{\underbar{26.03}}
&\scriptsize{\underbar{0.803}}
\\

\scriptsize{\textbf{DMPlug (ours)}} 

&\scriptsize{\textbf{0.073}}
&\scriptsize{\textbf{31.61}}
&\scriptsize{\textbf{0.882}}

&\scriptsize{\textbf{0.057}}
&\scriptsize{\textbf{32.83}}
&\scriptsize{\textbf{0.907}}

&\scriptsize{\textbf{0.083}}
&\scriptsize{\textbf{30.74}}
&\scriptsize{\textbf{0.882}}
\\
\scriptsize{\textbf{Ours vs. Best compe.}}
&\scriptsize{\textcolor{darkpastelgreen}{$-0.031$}}
&\scriptsize{\textcolor{darkpastelgreen}{$+3.09$}}
&\scriptsize{\textcolor{darkpastelgreen}{$+0.043$}}

&\scriptsize{\textcolor{darkpastelgreen}{$-0.047$}}
&\scriptsize{\textcolor{darkpastelgreen}{$+5.79$}}
&\scriptsize{\textcolor{darkpastelgreen}{$+0.073$}}

&\scriptsize{\textcolor{darkpastelgreen}{$-0.060$}}
&\scriptsize{\textcolor{darkpastelgreen}{$+4.71$}}
&\scriptsize{\textcolor{darkpastelgreen}{$+0.079$}}

\\

\hline

\hline
\end{tabular}
}
\end{center}
\vspace{-1em}
\end{table}

\subsection{IP tasks and experimental results} 
\label{subsec:exp_res}

\paragraph{Linear IPs} 
Due to space constraints, our experimental results on \textbf{\textcolor{umn_maroon}{Super-resolution \& inpainting}}  are included in \cref{app:quan_res}. Our DMPlug can lead the best SOTA methods by about \textcolor{darkpastelgreen}{$2$dB} in PSNR and \textcolor{darkpastelgreen}{$0.02$} in SSIM on average, respectively. 


\begin{wraptable}{r}{0.6\linewidth}
\vspace{-2em}
\centering
\caption{(\textcolor{red}{Nonlinear IP}) \textbf{BID with turbulence} with additive Gaussian noise ($\sigma = 0.01$). (\textbf{Bold}: best, \underbar{under}: second best, \textcolor{darkpastelgreen}{green}: performance increase, \textcolor{red}{red}: performance decrease)}
\label{tab:turbulence}
\setlength{\tabcolsep}{1.0mm}{
\begin{tabular}{c c c c c c c}
\hline

&\multicolumn{3}{c}{\scriptsize{\textbf{CelebA~\cite{liu_deep_2015} ($256 \times 256$)}}}
&\multicolumn{3}{c}{\scriptsize{\textbf{FFHQ~\cite{karras_style-based_2019} ($256 \times 256$)}}}
\\

\cmidrule(lr){2-4}
\cmidrule(lr){5-7}

&\scriptsize{LPIPS$\downarrow$}
&\scriptsize{PSNR$\uparrow$}
&\scriptsize{SSIM$\uparrow$}

&\scriptsize{LPIPS$\downarrow$}
&\scriptsize{PSNR$\uparrow$}
&\scriptsize{SSIM$\uparrow$}

\\
\hline

\scriptsize{DeBlurGANv2~\cite{kupyn_deblurgan-v2_2019}} 
&\scriptsize{0.356}
&\scriptsize{24.17}
&\scriptsize{0.739}

&\scriptsize{0.379}
&\scriptsize{23.52}
&\scriptsize{0.723}

\\

\scriptsize{Stripformer~\cite{tsai_stripformer_2022}} 
&\scriptsize{0.318}
&\scriptsize{24.97}
&\scriptsize{\underbar{0.745}}

&\scriptsize{0.343}
&\scriptsize{24.30}
&\scriptsize{\underbar{0.725}}
\\

\scriptsize{MPRNet~\cite{zamir_multi-stage_2021}} 
&\scriptsize{0.379}
&\scriptsize{22.64}
&\scriptsize{0.696}

&\scriptsize{0.399}
&\scriptsize{22.28}
&\scriptsize{0.682}
\\

\scriptsize{TSR-WGAN~\cite{jin_neutralizing_2021}}
&\scriptsize{0.304}
&\scriptsize{23.09}
&\scriptsize{0.732}

&\scriptsize{0.333}
&\scriptsize{22.56}
&\scriptsize{0.715}
\\

\scriptsize{ILVR~\cite{choi_ilvr_2021}} 
&\scriptsize{0.337}
&\scriptsize{21.25}
&\scriptsize{0.589}

&\scriptsize{0.375}
&\scriptsize{20.24}
&\scriptsize{0.554}
\\

\scriptsize{BlindDPS~\cite{chung_parallel_2022}} 
&\scriptsize{\textbf{0.137}}
&\scriptsize{\underbar{25.45}}
&\scriptsize{0.730}

&\scriptsize{\underbar{0.165}}
&\scriptsize{\underbar{24.40}}
&\scriptsize{0.712}
\\

\scriptsize{\textbf{DMPlug (ours)}} 
&\scriptsize{\underbar{0.146}}
&\scriptsize{\textbf{28.34}}
&\scriptsize{\textbf{0.790}}

&\scriptsize{\textbf{0.164}}
&\scriptsize{\textbf{27.91}}
&\scriptsize{\textbf{0.812}}
\\
\scriptsize{\textbf{Ours vs. Best compe.}}
&\scriptsize{\textcolor{red}{$-0.009$}}
&\scriptsize{\textcolor{darkpastelgreen}{$+2.89$}}
&\scriptsize{\textcolor{darkpastelgreen}{$+0.045$}}

&\scriptsize{\textcolor{darkpastelgreen}{$-0.001$}}
&\scriptsize{\textcolor{darkpastelgreen}{$+3.51$}}
&\scriptsize{\textcolor{darkpastelgreen}{$+0.087$}}

\\

\hline

\hline
\end{tabular}
}
\vspace{-1em}
\end{wraptable}

\paragraph{Nonlinear IPs} \textbf{\textcolor{umn_maroon}{Nonlinear deblurring}} 
We use the learned blurring operators from~\cite{tran_explore_2021} with a known Gaussian-shaped kernel and Gaussian additive noise with $\sigma=0.01$, following~\cite{song_solving_2023,chung_diffusion_2023}. 
We compare our DMPlug against several strong baselines: Blur Kernel Space (BKS)-styleGAN2~\cite{tran_explore_2021} based on GAN priors, BKS-generic~\cite{tran_explore_2021} based on Hyper-Laplacian priors~\cite{krishnan_fast_2009}, and DM-based methods that can handle nonlinear IPs, including MCG, ILVR, DPS and ReSample. Despite the advancements made by the recent ReSample~\cite{song_solving_2023} to enhance the DPS method, from \cref{tab:nonuniform}, it is evident that our DMPlug can still significantly outperform the SOTA ReSample by substantial margins. Specifically, our method improves LPIPS, PSNR, and SSIM by \textcolor{darkpastelgreen}{$0.05$}, \textcolor{darkpastelgreen}{$4.5$dB}, and \textcolor{darkpastelgreen}{$0.07$},  respectively, across the three datasets on average. In addition, our DMPlug delivers a more faithful and precise restoration of details, as shown in \cref{fig:ips}. 

\textbf{\textcolor{umn_maroon}{BID \& BID with turbulence}} BID is about recovering a sharp image $\mb x$ (and kernel $\mb k$) from $\mb y = \mb k \ast \mb x + \mb n$ where $\ast$ denotes the linear convolution and the spatially invariant blur kernel $\mb k$ is unknown; for BID with turbulence which often arises in long-range imaging through atmospheric turbulence, we model the forward process as a simplified ``tilt-then-blur'' process, following~\cite{chan_tilt-then-blur_2022}: $\mb y = \mb k \ast \mathcal{T}_{\mb \phi} \paren{\mb x} + \mb n$, where $\mathcal{T}_{\mb \phi}\paren{\cdot}$ is the tilt operation parameterized by unknown $\mb \phi$ (see more details in \cref{app:setup}). We mainly compare our DMPlug to two DM-based models, including ILVR and BlindDPS~\cite{chung_parallel_2022}, which is an extension of DPS. In addition, we choose two classical MAP-based methods, i.e., Pan-Dark Channel Prior (Pan-DCP)~\cite{pan_deblurring_2018} and Pan-$\ell_0$~\cite{pan_l_0_2017}, DIP-based SelfDeblur~\cite{ren_neural_2020}, and four methods that are based on supervised training on paired datasets, including DeBlurGANv2~\cite{kupyn_deblurgan-v2_2019}, Stripformer~\cite{tsai_stripformer_2022}, MPRNet~\cite{zamir_multi-stage_2021} and TSR-WGAN~\cite{jin_neutralizing_2021}. It is important to mention that \textbf{BlindDPS~\cite{chung_parallel_2022} use pretrained DMs not only for images but also for blur kernels (and tilt maps), lending it an unfair advantage over other methods}. Although our method is flexible in working with multiple DM priors, as shown in \cref{subsec:exp_ablation}, we opt to only use the pretrained DMs for images to ensure a fair comparison. \cref{tab:BID,tab:turbulence} show that our method, despite using fewer priors than BlindDPS, can surpass the best competing methods by approximately \textcolor{darkpastelgreen}{$0.03$}, \textcolor{darkpastelgreen}{$4.5$dB}, and \textcolor{darkpastelgreen}{$0.1$} in terms of LPIPS, PSNR, and SSIM, respectively, on average. In \cref{fig:ips}, the reconstructions of our method look sharper and more precise than those of the main competitors. 

\begin{table}[!htpb]
\centering 
\caption{(\textcolor{red}{Nonlinear IP}) \textbf{BID} with additive Gaussian noise ($\sigma = 0.01$). (\textbf{Bold}: best, \underbar{under}: second best, \textcolor{darkpastelgreen}{green}: performance increase, \textcolor{red}{red}: performance decrease)}
\vspace{-0.5em}
\label{tab:BID}
\setlength{\tabcolsep}{0.85mm}{
\begin{tabular}{c c c c c c c c c c c c c}
\hline

&\multicolumn{6}{c}{\scriptsize{\textbf{CelebA~\cite{liu_deep_2015} ($256 \times 256$)}}}
&\multicolumn{6}{c}{\scriptsize{\textbf{FFHQ~\cite{karras_style-based_2019} ($256 \times 256$)}}}

\\
\cmidrule(lr){2-7}
\cmidrule(lr){8-13}

&\multicolumn{3}{c}{\scriptsize{\textbf{Motion blur}}}
&\multicolumn{3}{c}{\scriptsize{\textbf{Gaussian blur}}}
&\multicolumn{3}{c}{\scriptsize{\textbf{Motion blur}}}
&\multicolumn{3}{c}{\scriptsize{\textbf{Gaussian blur}}}
\\

\cmidrule(lr){2-4}
\cmidrule(lr){5-7}
\cmidrule(lr){8-10}
\cmidrule(lr){11-13}

&\scriptsize{LPIPS$\downarrow$}
&\scriptsize{PSNR$\uparrow$}
&\scriptsize{SSIM$\uparrow$}

&\scriptsize{LPIPS$\downarrow$}
&\scriptsize{PSNR$\uparrow$}
&\scriptsize{SSIM$\uparrow$}

&\scriptsize{LPIPS$\downarrow$}
&\scriptsize{PSNR$\uparrow$}
&\scriptsize{SSIM$\uparrow$}

&\scriptsize{LPIPS$\downarrow$}
&\scriptsize{PSNR$\uparrow$}
&\scriptsize{SSIM$\uparrow$}
\\
\hline

\scriptsize{SelfDeblur~\cite{ren_neural_2020}} 
&\scriptsize{0.568}
&\scriptsize{16.59}
&\scriptsize{0.417}

&\scriptsize{0.579}
&\scriptsize{16.55}
&\scriptsize{0.423}

&\scriptsize{0.628}
&\scriptsize{16.33}
&\scriptsize{0.408}

&\scriptsize{0.604}
&\scriptsize{16.22}
&\scriptsize{0.410}
\\

\scriptsize{DeBlurGANv2~\cite{kupyn_deblurgan-v2_2019}} 
&\scriptsize{0.313}
&\scriptsize{20.56}
&\scriptsize{0.613}

&\scriptsize{0.350}
&\scriptsize{24.29}
&\scriptsize{0.743}

&\scriptsize{0.353}
&\scriptsize{19.67}
&\scriptsize{0.581}

&\scriptsize{0.374}
&\scriptsize{23.58}
&\scriptsize{0.726}
\\

\scriptsize{Stripformer~\cite{tsai_stripformer_2022}} 
&\scriptsize{0.287}
&\scriptsize{22.06}
&\scriptsize{0.644}

&\scriptsize{0.316}
&\scriptsize{25.03}
&\scriptsize{\underbar{0.747}}

&\scriptsize{0.324}
&\scriptsize{21.31}
&\scriptsize{0.613}

&\scriptsize{0.339}
&\scriptsize{\underbar{24.34}}
&\scriptsize{\underbar{0.728}}
\\

\scriptsize{MPRNet~\cite{zamir_multi-stage_2021}} 
&\scriptsize{0.332}
&\scriptsize{20.53}
&\scriptsize{0.620}

&\scriptsize{0.375}
&\scriptsize{22.72}
&\scriptsize{0.698}

&\scriptsize{0.373}
&\scriptsize{19.70}
&\scriptsize{0.590}

&\scriptsize{0.394}
&\scriptsize{22.33}
&\scriptsize{0.685}
\\

\scriptsize{Pan-DCP~\cite{pan_deblurring_2018}} 
&\scriptsize{0.606}
&\scriptsize{15.83}
&\scriptsize{0.483}

&\scriptsize{0.653}
&\scriptsize{20.57}
&\scriptsize{0.701}

&\scriptsize{0.616}
&\scriptsize{15.59}
&\scriptsize{0.464}

&\scriptsize{0.667}
&\scriptsize{20.69}
&\scriptsize{0.698}
\\

\scriptsize{Pan-$\ell_0$~\cite{pan_l_0_2017}} 
&\scriptsize{0.631}
&\scriptsize{15.16}
&\scriptsize{0.470}

&\scriptsize{0.654}
&\scriptsize{20.49}
&\scriptsize{0.675}

&\scriptsize{0.642}
&\scriptsize{14.43}
&\scriptsize{0.443}

&\scriptsize{0.669}
&\scriptsize{20.34}
&\scriptsize{0.671}
\\

\scriptsize{ILVR~\cite{choi_ilvr_2021}} 
&\scriptsize{0.398}
&\scriptsize{19.23}
&\scriptsize{0.520}

&\scriptsize{0.338}
&\scriptsize{21.20}
&\scriptsize{0.588}

&\scriptsize{0.445}
&\scriptsize{18.33}
&\scriptsize{0.484}

&\scriptsize{0.375}
&\scriptsize{20.45}
&\scriptsize{0.555}
\\

\scriptsize{BlindDPS~\cite{chung_parallel_2022}} 
&\scriptsize{\underbar{0.164}}
&\scriptsize{\underbar{23.60}}
&\scriptsize{\underbar{0.682}}

&\scriptsize{\underbar{0.173}}
&\scriptsize{\underbar{25.15}}
&\scriptsize{0.721}

&\scriptsize{\underbar{0.185}}
&\scriptsize{\underbar{21.77}}
&\scriptsize{\underbar{0.630}}

&\scriptsize{\underbar{0.193}}
&\scriptsize{23.83}
&\scriptsize{0.693}
\\

\scriptsize{\textbf{DMPlug (ours)}} 

&\scriptsize{\textbf{0.104}}
&\scriptsize{\textbf{29.61}}
&\scriptsize{\textbf{0.825}}

&\scriptsize{\textbf{0.140}}
&\scriptsize{\textbf{28.84}}
&\scriptsize{\textbf{0.795}}

&\scriptsize\textbf{{0.135}}
&\scriptsize{\textbf{27.99}}
&\scriptsize{\textbf{0.794}}

&\scriptsize{\textbf{0.169}}
&\scriptsize{\textbf{28.26}}
&\scriptsize{\textbf{0.811}}
\\
\scriptsize{\textbf{Ours vs. Best compe.}}
&\scriptsize{\textcolor{darkpastelgreen}{$-0.060$}}
&\scriptsize{\textcolor{darkpastelgreen}{$+6.01$}}
&\scriptsize{\textcolor{darkpastelgreen}{$+0.143$}}

&\scriptsize{\textcolor{darkpastelgreen}{$-0.033$}}
&\scriptsize{\textcolor{darkpastelgreen}{$+3.69$}}
&\scriptsize{\textcolor{darkpastelgreen}{$+0.048$}}

&\scriptsize{\textcolor{darkpastelgreen}{$-0.050$}}
&\scriptsize{\textcolor{darkpastelgreen}{$+6.22$}}
&\scriptsize{\textcolor{darkpastelgreen}{$+0.164$}}

&\scriptsize{\textcolor{darkpastelgreen}{$-0.024$}}
&\scriptsize{\textcolor{darkpastelgreen}{$+3.92$}}
&\scriptsize{\textcolor{darkpastelgreen}{$+0.083$}}

\\

\hline

\hline
\end{tabular}
}
\vspace{-1em}
\end{table}

\subsection{Robustness to unknown noise}
\label{subsec:exp_robust}

\begin{figure}[!htbp]
    \centering
    \vspace{-1em}
    \begin{minipage}{.7\textwidth}
        \centering
        \includegraphics[width=1\linewidth]{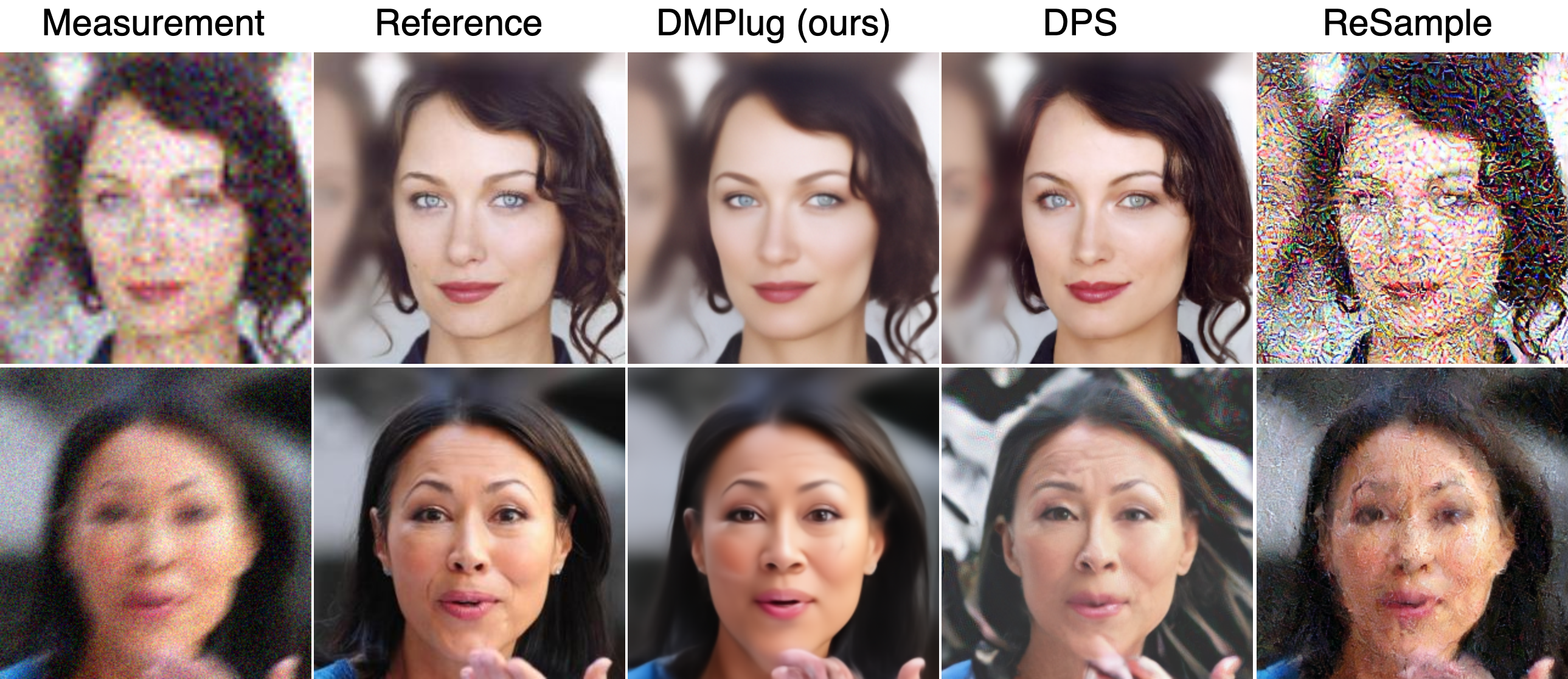}
    \end{minipage}
    \begin{minipage}{.25\textwidth}
        \centering
        \caption{(\textbf{Robustness}) Visualization of sample results from our DMPlug and main competing methods for $4 \times$ super-resolution (\textbf{top}) and nonlinear deblurring (\textbf{bottom}) with low-level Gaussian noise.}
        \label{fig:robsut} 
    \end{minipage}
    \vspace{-0.5em}
\end{figure}

For robustness experiments, we choose super-resolution and nonlinear deblurring to represent linear and nonlinear IPs, respectively. To simulate scenarios involving unknown noise, we generate measurements with four types of noise: Gaussian, impulse, shot, and speckle noise, and across two different noise levels: low (level-1) and high (level-2), following~\cite{hendrycks_benchmarking_2019} (see details in \cref{app:noise}), but we use the same formulation and code for each IP designed for mild Gaussian noise, regardless of the actual noise type and level. \cref{tab:robust_es} and \cref{fig:robsut} clearly show that (1) \textbf{most current IP solvers, except for DPS, suffer from the robustness issue}, corroborating the hypotheses made in \cref{sec:bg}, and (2) the peak performance of our DMPlug can lead SOTA methods by around \textcolor{darkpastelgreen}{$1$dB} and \textcolor{darkpastelgreen}{$3.5$dB} in PSNR for the two tasks, respectively. To check the compatibility of our method with ES-WMV~\cite{wang_early_2023}, we measure the detection performance via PSNR gaps, i.e., the absolute PSNR difference between the peak and the detected ES point following~\cite{wang_early_2023,li_deep_2023}. \cref{tab:robust_es} indicates that the detection gaps in the two exemplary tasks are nearly negligible, with PSNR gaps smaller than \textcolor{darkpastelgreen}{$0.5$dB} and \textcolor{darkpastelgreen}{$0.2$dB}, respectively. This suggests that \textbf{the proposed method is highly synergetic with ES-WMV}. 

\begin{table}[!htpb]
\caption{(\textbf{Robustness and ES}) \textbf{Super-resolution} and \textbf{nonlinear deblurring} on CelebA~\cite{liu_deep_2015} with different types and levels of noise. We only show PSNR$\uparrow$ and PSNR Gap$\downarrow$ to save space. (\textbf{Bold}: best, \underbar{under}: second best, \textcolor{darkpastelgreen}{green}: performance increase, \textcolor{red}{red}: performance decrease)}
\vspace{-1em}
\label{tab:robust_es}
\begin{center}
\setlength{\tabcolsep}{1.0mm}{
\begin{tabular}{c c c c c c c c c}
\hline

&\multicolumn{4}{c}{\scriptsize{\textbf{(\textcolor{red}{Linear}) Super-resolution ($4 \times$)}}}
&\multicolumn{4}{c}{\scriptsize{\textbf{(\textcolor{red}{Nonlinear}) Non-uniform image deblurring}}}

\\
\cmidrule(lr){2-5}
\cmidrule(lr){6-9}

&\multicolumn{1}{c}{\scriptsize{\textbf{Gaussian}}}
&\multicolumn{1}{c}{\scriptsize{\textbf{Impulse}}}
&\multicolumn{1}{c}{\scriptsize{\textbf{Shot}}}
&\multicolumn{1}{c}{\scriptsize{\textbf{Speckle}}}

&\multicolumn{1}{c}{\scriptsize{\textbf{Gaussian}}}
&\multicolumn{1}{c}{\scriptsize{\textbf{Impulse}}}
&\multicolumn{1}{c}{\scriptsize{\textbf{Shot}}}
&\multicolumn{1}{c}{\scriptsize{\textbf{Speckle}}}
\\

\cmidrule(lr){2-5}
\cmidrule(lr){6-9}

&\scriptsize{Low}/\scriptsize{High}

&\scriptsize{Low}/\scriptsize{High}

&\scriptsize{Low}/\scriptsize{High}

&\scriptsize{Low}/\scriptsize{High}

&\scriptsize{Low}/\scriptsize{High}

&\scriptsize{Low}/\scriptsize{High}

&\scriptsize{Low}/\scriptsize{High}

&\scriptsize{Low}/\scriptsize{High}

\\
\hline

\scriptsize{ADMM-PnP~\cite{chan_plug-and-play_2016}}
&\scriptsize{20.17}/\scriptsize{17.97}

&\scriptsize{14.28}/\scriptsize{14.52}

&\scriptsize{19.97}/\scriptsize{17.82}

&\scriptsize{19.42}/\scriptsize{18.41}

&\scriptsize{N/A}

&\scriptsize{N/A}

&\scriptsize{N/A}

&\scriptsize{N/A}

\\

\scriptsize{DMPS~\cite{meng_diffusion_2024}}
&\scriptsize{20.62}/\scriptsize{17.54}

&\scriptsize{18.78}/\scriptsize{16.05}

&\scriptsize{19.96}/\scriptsize{16.74}

&\scriptsize{20.77}/\scriptsize{18.73}

&\scriptsize{N/A}

&\scriptsize{N/A}

&\scriptsize{N/A}

&\scriptsize{N/A}
\\

\scriptsize{DDRM~\cite{kawar_denoising_2022}}
&\scriptsize{15.45}/\scriptsize{14.79}

&\scriptsize{14.82}/\scriptsize{14.14}

&\scriptsize{15.31}/\scriptsize{14.59}

&\scriptsize{15.46}/\scriptsize{15.03}

&\scriptsize{N/A}

&\scriptsize{N/A}

&\scriptsize{N/A}

&\scriptsize{N/A}
\\

\scriptsize{MCG~\cite{chung_improving_2022}}
&\scriptsize{17.43}/\scriptsize{15.83}

&\scriptsize{16.39}/\scriptsize{15.07}

&\scriptsize{17.19}/\scriptsize{15.49}

&\scriptsize{17.44}/\scriptsize{16.43}

&\scriptsize{12.88}/\scriptsize{12.85}

&\scriptsize{13.16}/\scriptsize{13.04}

&\scriptsize{13.21}/\scriptsize{13.13}

&\scriptsize{13.24}/\scriptsize{13.07}
\\

\scriptsize{ILVR~\cite{choi_ilvr_2021}}
&\scriptsize{21.08}/\scriptsize{21.03}

&\scriptsize{20.93}/\scriptsize{20.00}

&\scriptsize{21.19}/\scriptsize{21.12}

&\scriptsize{20.96}/\scriptsize{20.89}

&\scriptsize{21.70}/\scriptsize{21.43}

&\scriptsize{21.43}/\scriptsize{21.00}

&\scriptsize{21.56}/\scriptsize{21.24}

&\scriptsize{21.53}/\scriptsize{21.36}
\\

\scriptsize{DPS~\cite{chung_diffusion_2023}} 
&\scriptsize{\underbar{25.51}}/\scriptsize{\underbar{24.58}}

&\scriptsize{\underbar{24.89}}/\scriptsize{\underbar{23.92}}

&\scriptsize{\underbar{25.47}}/\scriptsize{\underbar{24.27}}

&\scriptsize{\underbar{25.69}}/\scriptsize{\underbar{24.97}}

&\scriptsize{\underbar{23.97}}/\scriptsize{\underbar{23.35}}

&\scriptsize{\underbar{23.74}}/\scriptsize{\underbar{23.18}}

&\scriptsize{\underbar{24.32}}/\scriptsize{\underbar{23.58}}

&\scriptsize{\underbar{23.45}}/\scriptsize{\underbar{23.61}}
\\

\scriptsize{ReSample~\cite{song_solving_2023}} 
&\scriptsize{14.30}/\scriptsize{13.04}

&\scriptsize{15.56}/\scriptsize{13.48}

&\scriptsize{14.38}/\scriptsize{12.87}

&\scriptsize{15.64}/\scriptsize{14.23}

&\scriptsize{23.17}/\scriptsize{20.45}

&\scriptsize{20.69}/\scriptsize{18.91}

&\scriptsize{22.94}/\scriptsize{20.11}

&\scriptsize{23.59}/\scriptsize{21.66}
\\

\scriptsize{BKS-styleGAN~\cite{tran_explore_2021}}
&\scriptsize{N/A}

&\scriptsize{N/A}

&\scriptsize{N/A}

&\scriptsize{N/A}

&\scriptsize{22.61}/\scriptsize{22.53}

&\scriptsize{22.64}/\scriptsize{22.34}

&\scriptsize{22.96}/\scriptsize{22.79}

&\scriptsize{22.70}/\scriptsize{22.56}
\\

\scriptsize{BKS-generic~\cite{tran_explore_2021}}
&\scriptsize{N/A}

&\scriptsize{N/A}

&\scriptsize{N/A}

&\scriptsize{N/A}

&\scriptsize{16.85}/\scriptsize{15.09}

&\scriptsize{14.86}/\scriptsize{13.44}

&\scriptsize{16.69}/\scriptsize{14.74}

&\scriptsize{17.04}/\scriptsize{15.99}
\\
\scriptsize{\textbf{DMPlug (ours)}} 
&\scriptsize{\textbf{26.49}}/\scriptsize{\textbf{25.29}}

&\scriptsize{\textbf{26.01}}/\scriptsize{\textbf{24.76}}

&\scriptsize{\textbf{26.34}}/\scriptsize{\textbf{26.34}}

&\scriptsize{\textbf{26.81}}/\scriptsize{\textbf{25.81}}

&\scriptsize{\textbf{27.58}}/\scriptsize{\textbf{26.60}}

&\scriptsize{\textbf{27.22}}/\scriptsize{\textbf{26.13}}

&\scriptsize{\textbf{27.71}}/\scriptsize{\textbf{26.55}}

&\scriptsize{\textbf{27.68}}/\scriptsize{\textbf{26.96}}
\\
\scriptsize{\textbf{Ours vs. Best compe.}}
&\scriptsize{\textcolor{darkpastelgreen}{$0.98$}}/\scriptsize{\textcolor{darkpastelgreen}{$0.71$}}

&\scriptsize{\textcolor{darkpastelgreen}{$1.12$}}/\scriptsize{\textcolor{darkpastelgreen}{$0.84$}}

&\scriptsize{\textcolor{darkpastelgreen}{$0.87$}}/\scriptsize{\textcolor{darkpastelgreen}{$2.07$}}

&\scriptsize{\textcolor{darkpastelgreen}{$1.12$}}/\scriptsize{\textcolor{darkpastelgreen}{$0.84$}}

&\scriptsize{\textcolor{darkpastelgreen}{$3.61$}}/\scriptsize{\textcolor{darkpastelgreen}{$3.25$}}

&\scriptsize{\textcolor{darkpastelgreen}{$3.48$}}/\scriptsize{\textcolor{darkpastelgreen}{$2.95$}}

&\scriptsize{\textcolor{darkpastelgreen}{$3.39$}}/\scriptsize{\textcolor{darkpastelgreen}{$2.97$}}

&\scriptsize{\textcolor{darkpastelgreen}{$4.23$}}/\scriptsize{\textcolor{darkpastelgreen}{$3.35$}}

\\
\scriptsize{\textbf{PSNR Gap$\downarrow$}}
&\scriptsize{0.36}/\scriptsize{0.46}

&\scriptsize{0.38}/\scriptsize{0.60}

&\scriptsize{0.25}/\scriptsize{0.49}

&\scriptsize{0.20}/\scriptsize{0.21}

&\scriptsize{0.15}/\scriptsize{0.12}

&\scriptsize{0.14}/\scriptsize{0.13}

&\scriptsize{0.10}/\scriptsize{0.19}

&\scriptsize{0.12}/\scriptsize{0.09}
\\

\hline

\hline
\end{tabular}
}
\end{center}
\vspace{-1em}
\end{table}

\subsection{Ablation studies}
\label{subsec:exp_ablation}
We conduct two ablation studies to demonstrate the flexibility of our method in terms of using multiple and non-DDIM DM priors and using alternative optimizers. \textbf{\textcolor{umn_maroon}{(Flexibility of using DM priors)}} First, we explore the possibility of using different types of DMs. For super-resolution, we show in \cref{tab:ab_prior} that latent diffusion models (LDMs)~\cite{vahdat2021score,rombach_high-resolution_2022} are also synergetic with our method. Next, we study the potential of using multiple DMs together, taking BID with turbulence as an example. Using extra pretrained DMs for blur kernels and tilt maps from~\cite{chung_parallel_2022}, our method can achieve even better reconstruction results. \textbf{\textcolor{umn_maroon}{(Flexibility of using alternative optimizers)}} Here, we test the built-in \textit{ADAM}~\cite{kingma_adam_2017} and \textit{L-BFGS}~\cite{liu_limited_1989} optimizers in \textit{PyTorch}, with several different learning rates to solve \cref{eq:ours}. As shown in \cref{tab:ab_opt}, the best \textit{ADAM} and \textit{L-BFGS} combinations can lead to comparable performance for our DMPlug. We choose \textit{ADAM} as the default optimizer because in \textit{PyTorch}, optimizing multiple groups of variables with different learning rates---as for the case of BID (with turbulence)---is easy to program with \textit{ADAM} but tricky for \textit{L-BFGS}.       

\begin{table}[!htbp]
    \centering
    \begin{minipage}[t]{0.52\textwidth}
        \centering
        \caption{\textbf{(Flexibility)} Ablation on different \textbf{priors for $\mb x$, $\mb k$ and $\mb \phi$} on $50$ cases from CelebA~\cite{liu_deep_2015} for super-resolution (SR), nonlinear deblurring (ND), and BID with turbulence.}
        \begin{tabular}{c c c c c}
            \hline
            &\scriptsize{Prior for $\mb x$} &
            \scriptsize{Prior for $\mb k$} & 
            \scriptsize{Prior for $\mb \phi$} & 
            \scriptsize{PSNR$\uparrow$}
            \\
            \hline

            \multirow{2}{*}{\rotatebox{90}{\scriptsize{SR}}}
            &\scriptsize{DM} &
            \scriptsize{N/A} & \scriptsize{N/A} & \scriptsize{{31.51}}
            \\
            &\scriptsize{LDM} &
            \scriptsize{N/A} & 
             \scriptsize{N/A} & \scriptsize{{31.10}}
             \\
            \hline

            \multirow{2}{*}{\rotatebox{90}{\scriptsize{ND}}}
            &\scriptsize{DM} &
            \scriptsize{N/A} & \scriptsize{N/A} & \scriptsize{{31.61}}
            \\
            &\scriptsize{LDM} &
            \scriptsize{N/A} & 
             \scriptsize{N/A} & \scriptsize{{30.64}}
             \\
            \hline
            
            \multirow{4}{*}{\rotatebox{90}{\scriptsize{Turbulence}}}
            &\scriptsize{DM} &
            \scriptsize{Simplex} & \scriptsize{None} & \scriptsize{28.21}
            \\
            &\scriptsize{DM} &
            \scriptsize{Simplex+DM} & 
             \scriptsize{None} & \scriptsize{{29.33}} 
             \\
             &\scriptsize{DM} &
            \scriptsize{Simplex} & \scriptsize{DM} & \scriptsize{28.35} 
            \\
            &\scriptsize{DM} &
            \scriptsize{Simplex+DM} & \scriptsize{DM} & \scriptsize{{28.91}} 
            \\
            \hline
        \end{tabular}

        \label{tab:ab_prior}
    \end{minipage}
    \hfill
    \begin{minipage}[t]{0.44\textwidth}
        \centering
        \caption{\textbf{(Flexibility)} Ablation on different \textbf{optimizers and learning rates} on $50$ cases from CelebA~\cite{liu_deep_2015} for super-resolution.}
        \begin{tabular}{c c c c c}
            \hline
            &
            \scriptsize{Learning rate} & 
            \scriptsize{LPIPS$\downarrow$} &
            \scriptsize{PSNR$\uparrow$} & 
            \scriptsize{SSIM$\uparrow$} 
             \\
            \hline

            \multirow{3}{*}{\rotatebox{90}{\scriptsize{ADAM}}}
            &\scriptsize{0.1} 
            & \scriptsize{0.375} 
            & \scriptsize{20.22} 
            & \scriptsize{0.479} \\
            &\scriptsize{0.01}
            & \scriptsize{{0.066}}
            & \scriptsize{{31.51}}
            & \scriptsize{{0.879}} \\
            &\scriptsize{0.001}
            & \scriptsize{{0.093}}
            & \scriptsize{31.45} 
            & \scriptsize{{0.879}} \\
            \hline

            \multirow{3}{*}{\rotatebox{90}{\scriptsize{L-BFGS}}}
            &\scriptsize{1.0} 
            & \scriptsize{0.254} 
            & \scriptsize{28.53} 
            & \scriptsize{0.786} \\
            &\scriptsize{0.1}
            & \scriptsize{0.173}
            & \scriptsize{{31.89}}
            & \scriptsize{{0.888}} \\
            &\scriptsize{0.01}
            & \scriptsize{0.225} 
            & \scriptsize{30.20} 
            & \scriptsize{0.839} \\
            \hline
        \end{tabular}
        
        \label{tab:ab_opt}
    \end{minipage}
\end{table}

\section{Discussion}
\label{sec:dis}
In this paper, we focus on solving IPs with pretrained DMs. To deal with \textbf{\textcolor{blue}{(Issue 1)}} insufficient manifold feasibility and \textbf{\textcolor{blue}{(Issue 2)}} insufficient measurement feasibility of the prevailing interleaving methods, we \textbf{pioneer a novel plug-in method}, DMPlug, and make it practical in terms of computation and memory requirements. Taking advantage of a benign ELTO property and integrating an ES method ES-WMV~\cite{wang_early_2023}, our method is \textbf{the first to achieve robustness to unknown noise \textcolor{blue}{(Issue 3)}}. Extensive experiment results demonstrate that our method can lead SOTA methods, both qualitatively and quantitatively---often by large margins, particularly for nonlinear IPs. As for limitations, our empirical results in \cref{subsec:formulation} suggest a fundamental gap between image generation and regression using pretrained DMs, that we have not managed to nail down. Also, our work is mostly empirical, and we leave a solid theoretical understanding for future work. 

\section*{Acknowledgements} 
Wang H. is partially supported by a UMN CSE DSI PhD Fellowship. Wan Y. is partially supported by a UMN CSE InterS\&Ections Seed Grant. This research is part of AI-CLIMATE:``AI Institute for Climate-Land Interactions, Mitigation, Adaptation, Tradeoffs and Economy,'' and is supported by USDA National Institute of Food and Agriculture (NIFA) and the National Science Foundation (NSF) National AI Research Institutes Competitive Award no. 2023-67021-39829. The authors acknowledge the Minnesota Supercomputing Institute (MSI) at the University of Minnesota for providing resources that contributed to the research results reported in this article. 

{\small 
\setlength{\bibsep}{0pt plus 0.3ex}
\bibliographystyle{IEEEtran}
\bibliography{ref}
} 

\newpage  
\appendix

\section{Remarks on the concurrent work~\cite{chihaoui_zero-shot_2023}}
\label{sec:remark_SHRED}
We became aware of the unpublished concurrent work, SHRED~\cite{chihaoui_zero-shot_2023}, when we were finalizing the details of our method in mid Jan 2024 (We started the current project in Oct 2023). Although the core idea of SHRED is the same as that of our DMPlug, there are a few crucial differences between \cite{chihaoui_zero-shot_2023} and the current paper. \textbf{(1) Motivation}: \cite{chihaoui_zero-shot_2023} aims only at addressing manifold feasibility, while the current paper targets manifold feasibility, measurement feasibility, and robustness to unknown noise, simultaneously;  \textbf{(2) Integral components}: Since achieving robustness to unknown noise is part of our goal, the ES-WMV ES strategy is integral to our method, besides the unified optimization formulation \cref{eq:ours} shared with \cite{chihaoui_zero-shot_2023}; \textbf{(3) Key hyperparameters}: To address the computational and memory bottleneck induced by reverse steps, we use only $3$ reverse steps for all IPs we test, vs. the $10$ or more reverse steps used in \cite{chihaoui_zero-shot_2023}. So, our current setting makes the method much more practical; \textbf{(4) Flexibility}: Our framework allows for the use of more than one pre-trained DM prior when available, as shown in \cref{subsec:exp_ablation}, while \cite{chihaoui_zero-shot_2023} uses only a single DM prior as proposed in their Eqs. (17) to (19); \textbf{(5) Experimental evaluation}: \cite{chihaoui_zero-shot_2023} focuses on linear IPs, including inpainting, super-resolution, compressive sensing, and one nonlinear IP---blind deconvolution, i.e., blind image deburring (BID). We focus our evaluation on nonlinear IPs, including nonlinear deblurring, BID, and BID with turbulence, besides linear IPs (inpainting and super-resolution). Moreover, \cite{chihaoui_zero-shot_2023} measures performance only by perceptual metrics, LPIPS~\cite{zhang_unreasonable_2018} and FID~\cite{heusel2017gans}, while we measure performance by both the perceptual metric LPIPS and the classical metrics PSNR and SSIM. In addition, \cite{chihaoui_zero-shot_2023} does not even compare their method with clear SOTA methods that use pre-trained DM, e.g., DPS ~\cite{chung_diffusion_2023}, which they obviously were aware of (cited in the paper), whereas our comparison is more comprehensive. 

\section{Spectral bias of our DMPlug}

\label{app:sb}

\begin{figure*}[!htbp]
    \centering 
    \includegraphics[width=0.95\linewidth]{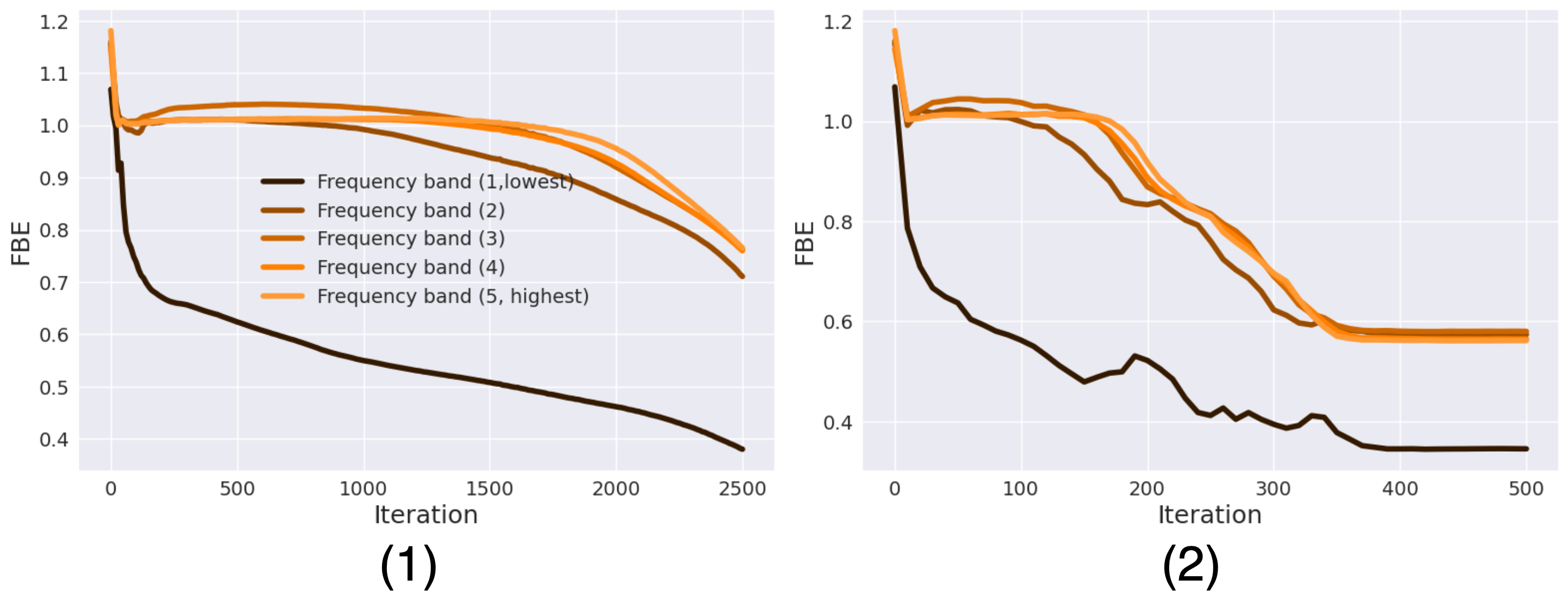}
    \caption{(1) The spectral bias of DMPlug with the \textit{ADAM} solver~\cite{kingma_adam_2017}. (2) The spectral bias of DMPlug with the \textit{L-BFGS} solver~\cite{liu_limited_1989}. The IP we experiment with here is regression against a noisy image with Gaussian noise at $\sigma = 0.08$, i.e., the same denoising problem as in \cref{fig:regression_denoising} (2). } 
    \label{fig:SB}
\end{figure*}

One may wonder why the ELTO phenomenon occurs. Here, we borrow the ideas of spectral biases and spectral analysis from the DIP literature~\cite{chakrabarty_spectral_2019,shi_measuring_2021,li_deep_2023,zhuang_blind_2024} to shed some light on this. The theory of spectral biases for DIP states that low-frequency components are learned much faster than high-frequency components during DIP learning. \textbf{Spectral biases in DIP lead to the ELTO phenomenon in DIP}: because natural images are typically low-frequency dominant~\cite{ruderman_statistics_1994}, the different learning paces imply that DIP learns mostly the desired image content (low-frequency image content plus the low-frequency part of noise) in the early stage, but gradually picks up the high-frequency part of noise in the late stage, resulting in performance degradation after a certain quality peak. Spectral analysis provides a quantitative visualization of spectral biases. 

Here, we perform a similar spectral analysis of our DMPlug learning process to demonstrate its spectral biases, which causes the ELTO phenomenon. To measure spectral biases, we follow~\cite{li_deep_2023,zhuang_blind_2024} and use \textit{frequency band errors (FBEs)}. For an groundtruth image $\mb x$ and its estimate $\wh{\mb x}$, the calculation of this metric goes as follows. First, we calculate the pointwise relative error pointwise in the Fourier domain, i.e., $\abs*{\mc F(\mb x) - \mc F(\wh{\mb x})}/\abs*{\mc F(\mb x)}$. Then, we divide the Fourier frequencies into five radial bands, compute the bandwise mean errors, i.e., the frequency-band errors (FBEs). 
 
We visualize the evolution of FBEs of DMPlug over all five frequency bands in \cref{fig:SB}. The disparate learning paces across the frequency bands are evident: the lowest-frequency band is learned much more rapidly than the other bands, which is consistent between two different optimization solvers. With spectral biases similar to those in DIP, we can explain the ELTO phenomenon in DMPlug following the argument above for DIP. 

\section{Setup details of the inverse problems we test}
\label{app:setup}

For super-resolution, inpainting, and nonlinear deblurring, we use the forward models from two of main competing methods~\cite{chung_diffusion_2023,song_solving_2023}; for BID and BID with turbulence, we follow the forward models from BlindDPS~\cite{chung_parallel_2022}---the SOTA DM-based method. Moreover, following~\cite{chung_diffusion_2023,song_solving_2023}, all the measurements contain additive Gaussian noise with $\sigma = 0.01$. Our loss $\ell$ by default is the MSE loss. 

\subsection{Super-resolution}
\label{app:sr}

In noisy image super-resolution, the goal is to reconstruct a clean RGB image $\mb x$ from a noisy downsampled version $\mb y = \mc{D} \paren{\mb x} + \mb n$, where $\mc{D}\paren{\cdot}: [0, 1]^{3 \times tH \times tW} \rightarrow [0, 1]^{3 \times H \times W}$ is a downsampling operator that resizes an image with dimensions $tH$ by $tW$ by the factor $t$ and $\mb n$ models additive noise. We set $t = 4$ in \cref{sec:exps}. To ensure a fair comparison, we do not include any explicit regularization terms in the formulation: 
\begin{align} \label{eq:sr}
    (\text{\textbf{Super-resolution}}) \; \mb z^* = \argmin\nolimits_{\mb z}\; \ell\paren{\mb y, \mc{D}\paren{\mc R\paren{\mb z}}}, \quad \quad \mb x^* = \mc{R}\paren{\mb z^*}. 
\end{align}

\subsection{Inpainting}
\label{app:inp}

In noisy image inpainting, a clean RGB image $\mb x \in [0, 1]^{3 \times H \times W}$ is only partially observed and then contaminated by additive noise $\mb n$, described by the forward model $\mb y = \mb m \odot \mb x + \mb n$, where $\mb m \in \{0, 1\}^{3 \times H \times W}$ is a binary mask and $\odot$ denotes the Hadamard product. Given $\mb y$ and $\mb m$, the goal is to reconstruct $\mb x$. Following~\cite{song_solving_2023}, the masks for the three channels of $\mb m$ are identical, and $70\%$ of the mask values are randomly set to $0$. To ensure a fair comparison, we do not include any explicit regularization terms in the formulation: 
\begin{align} \label{eq:inp}
    (\text{\textbf{Inpainting}}) \; \mb z^* = \argmin\nolimits_{\mb z}\; \ell\paren{\mb y, \mb m \odot \mc R\paren{\mb z}}, \quad \quad \mb x^* = \mc{R}\paren{\mb z^*}.
\end{align}

\subsection{Nonlinear deblurring}
\label{app:nonuniform}
We follow the setup in \cite{chung_diffusion_2023} which is inspired by ~\cite{tran_explore_2021}. Recently, \cite{tran_explore_2021} has proposed to learn data-driven blurring models from paired blurry-sharp training sets of the form $\{(\mb y_i, \mb x_i)\}_{i=1, \dots, N}$ through 
\begin{align} \label{eq:train_non_uniform_forward}
   \mb \alpha^*,  \mb \beta^* = \argmin\nolimits_{\mb \alpha, \mb \beta} \sum_{i=1}^{N} || \mb y_i - \mc F_{\mb \alpha} \paren{\mb x_i, \mc{G}_{\mb \beta}\paren{\mb x_i, \mb y_i}} ||,
\end{align}
where $\mc{G}_{\mb \beta}\paren{\cdot,\cdot}$ predicts the latent blur kernel associated with the input blurry-sharp image pair, and $\psi_{\mb \beta}\paren{\cdot,\cdot}$ models real-world nonlinear blurring process given the input image-kernel pair. To study the performance of its DPS method on nonlinear IPs, \cite{chung_diffusion_2023} proposes the following nonlinear deblurring problem with a known Gaussian-shaped kernel: 
\begin{align}
    \mb y = \mc F_{\mb \alpha^*} (\mb x, \mb g) + \mb n, \quad \text{where} \; \mb g \in \bb R^{64 \times 64}\, \text{is Gaussian-shaped with $\sigma = 3.0$}. 
\end{align}
The task here is to recover $\mb x$ from $\mb y$ and the forward model $\mc F_{\mb \alpha^*}(\cdot, \mb g)$. Our formulation follows~\cite{chung_diffusion_2023} and does not include extra regularizers: 
\begin{align} \label{eq:non_uniform_our}
    (\text{\textbf{Nonlinear deblurring}}) \; \mb z^* = \argmin\nolimits_{\mb z}\; \ell\paren{\mb y, \mc F_{\mb \alpha^*}\paren{\mc R\paren{\mb z}, \mb g}}, \quad \quad \mb x^* = \mc{R}\paren{\mb z^*}.
\end{align} 

\subsection{Blind image deblurring (BID)}
\label{app:bid}
BID is about recovering a sharp image $\mb x$ from $\mb y = \mb k \ast \mb x + \mb n$ where $\ast$ denotes the linear convolution and the spatially invariant blur kernel $\mb k$ is also unknown. It is a nonlinear IP because the forward model $\mc A (\mb k, \mb x) = \mb k \ast \mb x$ is nonlinear. In solving BID under the regularized data-fitting framework in \cref{eq:MAP}, $\ell_2$ or $\ell_1$ data-fitting loss, and the sparse gradient prior to $\mb x$ enforced by $R_{\mb x}\paren{\mb x} = \norm{\nabla \mb x}_1$ or $R_{\mb x}\paren{\mb x} = \norm{\nabla \mb x}_1/\norm{\nabla \mb x}_2$ are typically used. In addition, because of the scale ambiguity in the forward model, i.e., $\mb k \ast \mb x = (\alpha \mb k) \ast (\frac{1}{\alpha} \mb x)$ for any $\alpha > 0$, the scale of $\mb k$ is often fixed by requiring $\mb k$ to be on the standard \textbf{simplex} (i.e., $\mb k \ge \mb 0, \mb 1^\T \mb k = 1$) or on the sphere (i.e., $\norm{\mb k}_2 = 1$). \cite{koh_single-image_2021,zhang_deep_2022,zhuang_blind_2024} provide detailed coverage of these priors and regularizers. For our experiments here, we follow the settings in BlindDPS~\cite{chung_diffusion_2023}: all the blur kernels are simulated with the size of $64 \times 64$; the standard deviation of the Gaussian kernels is set to $3.0$ and the intensity of the motion blur kernels is adjusted to $0.5$. It is important to mention that \textbf{BlindDPS~\cite{chung_parallel_2022} uses pretrained DMs not only for images but also for blur kernels, giving it an unfair advantage over other methods}. But to ensure a fair comparison with other competing methods that only use data-driven image priors, we only use pretrained DMs for the image, plus the typical simplex constraint that is also used in BlindDPS: 
\begin{align} \label{eq:MAP_BID} 
(\text{\textbf{BID}}) \; \mb z^*, \mb k^* = \argmin\nolimits_{\mb z, \mb k}\; \ell\paren{\mb y, \mathrm{SoftMax}\paren{\mb k} \ast \mc R\paren{\mb z}}, \quad \quad \mb x^* = \mc{R}\paren{\mb z^*},
\end{align}
where $\mathrm{SoftMax}(\mb k)$ leads to a kernel estimate that lies on the standard simplex.  

\subsection{BID with turbulence}
\label{app:tur}
BID with turbulence often arises in long-range imaging through atmospheric turbulence. The forward imaging process can be modeled as a simplified ``tilt-then-blur'' process, following~\cite{chan_tilt-then-blur_2022}: $\mb y = \mb k \ast \mathcal{T}_{\mb \phi} \paren{\mb x} + \mb n$, where the tilt operator $\mathcal{T}_{\mb \phi}(\cdot)$ applies the spatially varying vector field $\mb \phi$ to the image $\mb x$ so that pixels of $\mb x$ are moved around according to the vector field $\mb \phi$, i.e., $\mathcal{T}_{\mb \phi} \paren{\mb x}[\mb p_i + \mb \phi_i] = \mb x[\mb p_i]$ where $\mb p_i$'s are the pixel coordinates. In BID with turbulence, none of the $\mb k, \mb x, \mb \phi$ is known and the task is to jointly estimate them from the measurement $\mb y$. So, it is clear that this BID variant is strictly more difficult than BID itself. We follow BlindDPS~\cite{chung_parallel_2022} for data generation: the blur kernel comes from 
the point spread function (PSF) and takes a Gaussian shape with standard deviation $3.0$; the tilt maps are generated as iid Gaussian random vectors over the pixel grid. Similar to the BID case, 
\textbf{BlindDPS~\cite{chung_parallel_2022} use pretrained DMs for all three objects: image, kernel, and tilt map, unfair to other methods}. We again use pretrained DMs for the image only. For the kernel, we again only impose the simplex constraint through a $\mathrm{SoftMax}$ activation. For the tilt map, inspired by the fact that the tilt vectors are small-magnitude random vectors, we initialize the map from a zero-mean Gaussian distribution with a small standard deviation and set an extremely small learning rate. Our final formulation for this task is 
{\small 
\begin{align} \label{eq:MAP_tur}
(\text{\textbf{BID with turbulence}}) \; \mb z^*, \mb k^*, \mb \phi^*  = \argmin\nolimits_{\mb z, \mb k, \mb \phi}\; \ell\paren{\mb y, \mathrm{SoftMax}\paren{\mb k} \ast \mathcal{T}_{\mb \phi}\paren{\mc R\paren{\mb z}}}, \; \mb x^* = \mc{R}\paren{\mb z^*}. 
\end{align}  
} 

\section{More implementation details}
\label{app:imp}

\subsection{Noise generation}
\label{app:noise}

Following \cite{hendrycks_benchmarking_2019}\footnote{\url{https://github.com/hendrycks/robustness}}, we simulate four types of noise, with two intensity levels for each type. The detailed information is as follows. \textbf{Gaussian noise:} zero-mean additive Gaussian noise with variance $0.08$ and $0.12$ for low and high noise levels, respectively; \textbf{Impulse noise:} also known as salt-and-pepper noise, replacing each pixel with probability $p \in [0, 1]$ in a white or black pixel with half chance each. Low and high noise levels correspond to $p = 0.03$ and $0.06$, respectively; \textbf{Shot noise:} also known as Poisson noise. For each pixel, $x \in [0, 1]$, the noisy pixel is Poisson distributed with the rate $\lambda x$, where $\lambda$ is $60$ and $25$ for low and high noise levels, respectively; \textbf{Speckle noise:} for each pixel $x \in [0, 1]$, the noisy pixel is $x\paren{1+\epsilon}$, where $\epsilon$ is zero-mean Gaussian with a variance level $0.15$ and $0.20$ for low and high noise levels, respectively.

\subsection{Additional implementation details of our method}
\label{app:imp_ours} 

We employ the following setup for our methods across all IP tasks. For $\mc{R}\paren{\cdot}$, we take the standard pretrained DMs from~\cite{dhariwal_diffusion_2021}\footnote{\url{https://github.com/openai/guided-diffusion}} and~\cite{choi_ilvr_2021}\footnote{\url{https://github.com/jychoi118/ilvr_adm?tab=readme-ov-file}} and use the standard DDIM~\cite{song_denoising_2022} sampler with only $3$ reverse steps based on \cref{fig:psnrvstime1}; we also use pretrained latent diffusion models (LDMs)~\cite{rombach_high-resolution_2022}\footnote{\url{https://github.com/CompVis/latent-diffusion}} to obtain the results reported in \cref{tab:ab_prior}. We use the pretrained DMs for blur kernels and tilt maps from~\cite{chung_parallel_2022}\footnote{\url{https://github.com/BlindDPS/blind-dps}} to obtain the results reported in \cref{tab:ab_prior}. For $\ell\paren{\cdot}$, we choose the standard \textit{MSE} loss. For $\Omega\paren{\cdot}$, we use the typical explicit regularizers for each task to make the comparisons fair. The default optimizer is \textit{ADAM} and the learning rate (LR) for $\mb z$ is $\num{1e-2}$; for BID (with turbulence), the LRs for blur kernel and the tilt map are $\num{1e-1}$ and $\num{1e-7}$, respectively. For the maximum numbers of iterations, we set $5,000$ and $10,000$ for linear and nonlinear IPs, respectively, which empirically allow good convergence. We perform all experiments on \textit{NVIDIA A100} GPUs with $40$GB memory each.

\begin{figure}[!htbp]
  \centering
  \begin{minipage}{.8\linewidth}
\begin{algorithm}[H]
\caption{DMPlug+ES--WMV for solving general IPs}
\label{alg:es} 
\begin{algorithmic}[1]
\Require \# diffusion steps $T$, $\mb y$, window size $W$, patience $P$, empty queue $\mc Q$, iteration counter $e = 0$, $\mathrm{VAR}_{\min} = \infty$
\While{not stopped}
\For{$i = T - 1$ to $0$} \tikzmark{top}
\State $\hat{\mb s} \gets \mb \varepsilon_{\mb \theta}^{(i)}\paren{\mb z_i^e}$
\State $\hat{\mb z}_0^e \gets \frac{1}{\sqrt{\Bar{\alpha}_i}}\paren{\mb z_i^e - \sqrt{1 - \Bar{\alpha}_i} \hat{\mb s}}$
\State $\mb z_{i-1}^e \gets$ DDIM reverse with $\hat{\mb z}_0^e, \hat{\mb s}$ \tikzmark{right}
\EndFor \tikzmark{bottom} 
\State Update $\mb z_T^{e+1}$ from $\mb z_T^{e}$ via a gradient update for \cref{eq:ours}
\State push $\mc R\paren{\mb z_T^{e+1}}$ to $\mc Q$, pop queue if $\abs{\mc Q} > W$
\If{$\abs{\mc Q} = W$}
\State compute $\mathrm{VAR}$ of elements in $\mc Q$ via \cref{eq:var}
\If{$\mathrm{VAR} < \mathrm{VAR}_{\min}$}
\State $\mathrm{VAR}_{\min} \leftarrow \mathrm{VAR}$, $\mb z^{*} \leftarrow \mb z_T^{e+1}$
\EndIf
\If{$\mathrm{VAR}_{\min}$ stagnates for $P$ iterations}
\State stop and return $\mb z^{*}$ 
\EndIf
\EndIf
\State $e = e+1$
\EndWhile
\Ensure Recovered object $\mc R\paren{\mb z^{*}}$
\end{algorithmic}
\end{algorithm}
  \end{minipage}
\end{figure}

\textbf{\textcolor{umn_maroon}{Early stopping (ES) using ES-WMV}} 
The ``early-learning-then-overfitting'' (\textbf{ELTO}) phenomenon has been widely reported in the literature on deep image prior (DIP)~\cite{ulyanov_deep_2020,li_self-validation_2021,li_deep_2023,shi_measuring_2021,wang_early_2023}, and ES-WMV~\cite{wang_early_2023} is an ES strategy that achieves the SOTA ES performance for DIP applied to various IPs. In ES-EMV, the running variance (VAR) is defined as: 
\begin{align}  \label{eq:var}
    \mathrm{VAR}\paren{t} \doteq \frac{1}{W}\sum_{w=0}^{W-1} \|\mb x^{t+w} - 1/W \cdot \sum_{i=0}^{W-1} \mb x^{t+i}\|_F^2, 
\end{align}
where $W$ is the window size and $\mb x^i$ denotes the recovery at iteration $i$. \cite{wang_early_2023} observes that \textbf{the first major valley} of the VAR curve is often well aligned with the peak of the PSNR curve. Based on this, \cite{wang_early_2023} introduces an online algorithm to detect the first major valley of the VAR curve: if the minimal VAR does not change over $P$ consecutive steps, i.e., the VAR does not decrease further over $P$ consecutive steps, the iteration process is stopped. The combined algorithm, ES-WMV-integrated DMPlug, is described in \cref{alg:es}. For implementation, we use the official code of ES-WMV\footnote{\url{https://github.com/sun-umn/Early_Stopping_for_DIP}}. For super-resolution with unknown noise, we set its patience number as $100$ and its window size as $10$; for nonlinear deblurring with unknown noise, we set its patience number as $300$ and its window size as $50$. \cref{tab:robust_es} indicates that the detection gaps in the two exemplary tasks are nearly negligible, with PSNR gaps smaller than \textcolor{darkpastelgreen}{$0.5$dB} and \textcolor{darkpastelgreen}{$0.2$dB}, respectively. This suggests that \textbf{our DMPlug is highly synergetic with ES-WMV}.

\subsection{Implementations of competing methods}
\label{app:imp_comp}

We use the default code and settings of each competitor's official implementation, listed below. 

 {\small
\begin{itemize} 
    \item ADMM-PnP~\cite{chan_plug-and-play_2016}: \url{https://github.com/kanglin755/plug_and_play_admm}
    \item DMPS~\cite{meng_diffusion_2024}: \url{https://github.com/mengxiangming/dmps}
    \item DDRM~\cite{kawar_denoising_2022}: \url{https://github.com/bahjat-kawar/ddrm}
    \item DPS~\cite{chung_diffusion_2023} \& MCG~\cite{chung_improving_2022}: \url{https://github.com/DPS2022/diffusion-posterior-sampling}
    \item ILVR~\cite{choi_ilvr_2021}: \url{https://github.com/jychoi118/ilvr_adm}
    \item ReSample~\cite{song_solving_2023}: \url{https://github.com/soominkwon/resample/tree/main}
    \item BKS~\cite{tran_explore_2021}: \url{https://github.com/VinAIResearch/blur-kernel-space-exploring}
    \item SelfDeblur~\cite{ren_neural_2020}: \url{https://github.com/csdwren/SelfDeblur}
    \item DeBlurGANv2~\cite{kupyn_deblurgan-v2_2019}: \url{https://github.com/VITA-Group/DeblurGANv2}
    \item Stripformer~\cite{tsai_stripformer_2022}: \url{https://github.com/pp00704831/Stripformer-ECCV-2022-}
    \item MPRNet~\cite{zamir_multi-stage_2021}: \url{https://github.com/swz30/MPRNet}
    \item Pan-DCP~\cite{pan_deblurring_2018}: \url{https://jspan.github.io/projects/dark-channel-deblur/index.html}
    \item Pan-$\ell_0$~\cite{pan_l_0_2017}: \url{https://jspan.github.io/projects/text-deblurring/index.html}
    \item BlindDPS~\cite{chung_parallel_2022}: \url{https://github.com/BlindDPS/blind-dps}
    \item TSR-WGAN~\cite{jin_neutralizing_2021}: \url{https://codeocean.com/capsule/9958894/tree/v1}
    \item FPS~\cite{dou_diffusion_2023}: \url{https://github.com/ZehaoDou-official/FPS-SMC-2023}
    \item DiffPIR~\cite{zhu_denoising_2023}: \url{https://github.com/yuanzhi-zhu/DiffPIR}
\end{itemize} 
 }

\section{More experiment results}
\label{app:more_res}

\subsection{Computational Efficiency}
\label{app:costs}

As shown in \cref{tab:costs}, we find that the memory usage of our method is in the same order as that of most other algorithms. When it comes to speed, our implementation using L-BFGS is significantly faster than the ADAM version. Although our method with L-BFGS remains somewhat slow, it surpasses the performance of ReSample. We will explore further acceleration of our method in future work.

\begin{table}[!htpb]
\scriptsize
\caption{Wall-clock time (seconds) and memory usage (GB) of various algorithms for \textbf{super-resolution} on the CelebA~\cite{liu_deep_2015} dataset, tested on a single \textit{NVIDIA A100 GPU}. (\textbf{Ours-A}: ours with ADAM, \textbf{Ours-L}: ours with L-BFGS)}
\label{tab:costs}
\begin{tabular}{llllllllllll}
\hline
                     & \textbf{ADMM-PnP} & \textbf{DMPS} & \textbf{DDRM} & \textbf{MCG} & \textbf{DPS} & \textbf{DDNM} & \textbf{FPS} & \textbf{DiffPIR} & \textbf{ReSample} & \textbf{Ours-A} & \textbf{Ours-L} \\
                     \hline
Time    & 6                 & 42            & 30            & 43           & 43           & 14            & 62           & 4                & 367               & 635                  & 255                    \\
Memory & 0.42              & 5.10          & 4.99          & 2.80         & 2.79         & 5.11          & 20.63        & 1.44             & 4.87              & 6.59                 & 6.74          
\\
\hline
\end{tabular}
\end{table}

\subsection{Quantitative results}
\label{app:quan_res}

\textbf{\textcolor{umn_maroon}{Super-resolution \& inpainting}} Following~\cite{song_solving_2023,chung_diffusion_2023}, to generate measurements, we use \textit{bicubic} downsampling for super-resolution ($4 \times$) and a random mask with $70\%$ missing pixels for inpainting; all measurements contain additive Gaussian noise with $\sigma = 0.01$. We compare our method with Plug-and-Play using ADMM (ADMM-PnP)~\cite{chan_plug-and-play_2016} and several SOTA DM-based methods: Diffusion Model based Posterior Sampling (DMPS)~\cite{meng_diffusion_2024}, Denoising Diffusion Destoration Models (DDRM)~\cite{kawar_denoising_2022}, Manifold Constrained Gradients (MCG)~\cite{chung_improving_2022}, Iterative Latent Variable Refinement (ILVR)~\cite{choi_ilvr_2021}, Diffusion Posterior Sampling (DPS)~\cite{chung_diffusion_2023}, and ReSample~\cite{song_solving_2023}. The quantitative results are reported in \cref{tab:sr_inp,tab:sr_inp_bedroom,tab:sr_inp_more}, with qualitative results visualized in \cref{fig:ips}. It is clear that our DMPlug consistently outperforms all competing methods on the three datasets, both quantitatively and qualitatively. Specifically, while there is no significant improvements in terms of LPIPS, the proposed method can lead the best SOTA methods by about \textcolor{darkpastelgreen}{$2$dB} in PSNR and \textcolor{darkpastelgreen}{$0.02$} in SSIM on average, respectively. 

\begin{table}[!htpb]
\caption{(\textcolor{red}{Linear IPs}) \textbf{Super-resolution} and \textbf{inpainting} with additive Gaussian noise ($\sigma = 0.01$). (\textbf{Bold}: best, \underbar{under}: second best, \textcolor{darkpastelgreen}{green}: performance increase, \textcolor{red}{red}: performance decrease)}
\vspace{-1.5em}
\label{tab:sr_inp}
\begin{center}
\setlength{\tabcolsep}{0.85mm}{
\begin{tabular}{c c c c c c c c c c c c c}
\hline

&\multicolumn{6}{c}{\scriptsize{\textbf{Super-resolution ($4 \times$)}}}
&\multicolumn{6}{c}{\scriptsize{\textbf{Inpainting (Random $70\%$)}}}

\\
\cmidrule(lr){2-7}
\cmidrule(lr){8-13}

&\multicolumn{3}{c}{\scriptsize{\textbf{CelebA~\cite{liu_deep_2015} ($256 \times 256$)}}}
&\multicolumn{3}{c}{\scriptsize{\textbf{FFHQ~\cite{karras_style-based_2019} ($256 \times 256$)}}}
&\multicolumn{3}{c}{\scriptsize{\textbf{CelebA~\cite{liu_deep_2015} ($256 \times 256$)}}}
&\multicolumn{3}{c}{\scriptsize{\textbf{FFHQ~\cite{karras_style-based_2019} ($256 \times 256$)}}}
\\

\cmidrule(lr){2-4}
\cmidrule(lr){5-7}
\cmidrule(lr){8-10}
\cmidrule(lr){11-13}

&\scriptsize{LPIPS$\downarrow$}
&\scriptsize{PSNR$\uparrow$}
&\scriptsize{SSIM$\uparrow$}

&\scriptsize{LPIPS$\downarrow$}
&\scriptsize{PSNR$\uparrow$}
&\scriptsize{SSIM$\uparrow$}

&\scriptsize{LPIPS$\downarrow$}
&\scriptsize{PSNR$\uparrow$}
&\scriptsize{SSIM$\uparrow$}

&\scriptsize{LPIPS$\downarrow$}
&\scriptsize{PSNR$\uparrow$}
&\scriptsize{SSIM$\uparrow$}
\\
\hline

\scriptsize{ADMM-PnP~\cite{chan_plug-and-play_2016}}
&\scriptsize{0.217}
&\scriptsize{26.99}
&\scriptsize{0.808}

&\scriptsize{0.229}
&\scriptsize{26.25}
&\scriptsize{0.794}

&\scriptsize{0.091}
&\scriptsize{31.94}
&\scriptsize{0.923}

&\scriptsize{0.104}
&\scriptsize{30.64}
&\scriptsize{0.901}
\\

\scriptsize{DMPS~\cite{meng_diffusion_2024}}
&\scriptsize{\underbar{0.070}}
&\scriptsize{\underbar{28.89}}
&\scriptsize{\underbar{0.848}}

&\scriptsize{\textbf{0.076}}
&\scriptsize{\underbar{28.03}}
&\scriptsize{\underbar{0.843}}

&\scriptsize{0.297}
&\scriptsize{24.52}
&\scriptsize{0.693}

&\scriptsize{0.326}
&\scriptsize{23.31}
&\scriptsize{0.664}
\\

\scriptsize{DDRM~\cite{kawar_denoising_2022}}
&\scriptsize{0.226}
&\scriptsize{26.34}
&\scriptsize{0.754}

&\scriptsize{0.282}
&\scriptsize{25.11}
&\scriptsize{0.731}

&\scriptsize{0.185}
&\scriptsize{26.10}
&\scriptsize{0.712}

&\scriptsize{0.201}
&\scriptsize{25.44}
&\scriptsize{0.722}
\\

\scriptsize{MCG~\cite{chung_improving_2022}}
&\scriptsize{0.725}
&\scriptsize{19.88}
&\scriptsize{0.323}

&\scriptsize{0.786}
&\scriptsize{18.20}
&\scriptsize{0.271}

&\scriptsize{1.283}
&\scriptsize{10.16}
&\scriptsize{0.049}

&\scriptsize{1.276}
&\scriptsize{10.37}
&\scriptsize{0.050}
\\

\scriptsize{ILVR~\cite{choi_ilvr_2021}}
&\scriptsize{0.322}
&\scriptsize{21.63}
&\scriptsize{0.603}

&\scriptsize{0.360}
&\scriptsize{20.73}
&\scriptsize{0.570}

&\scriptsize{0.447}
&\scriptsize{15.82}
&\scriptsize{0.484}

&\scriptsize{0.483}
&\scriptsize{15.10}
&\scriptsize{0.450}
\\

\scriptsize{DPS~\cite{chung_diffusion_2023}} 
&\scriptsize{0.087}
&\scriptsize{28.32}
&\scriptsize{0.823}

&\scriptsize{0.098}
&\scriptsize{27.44}
&\scriptsize{0.814}

&\scriptsize{0.043}
&\scriptsize{\underbar{32.24}}
&\scriptsize{\underbar{0.924}}

&\scriptsize{0.046}
&\scriptsize{\underbar{30.95}}
&\scriptsize{\underbar{0.913}}
\\

\scriptsize{ReSample~\cite{song_solving_2023}} 
&\scriptsize{0.080}
&\scriptsize{28.29}
&\scriptsize{0.819}

&\scriptsize{0.108}
&\scriptsize{25.22}
&\scriptsize{0.773}

&\scriptsize{\textbf{0.039}}
&\scriptsize{30.12}
&\scriptsize{0.904}

&\scriptsize{\underbar{0.044}}
&\scriptsize{27.91}
&\scriptsize{0.884}
\\

\scriptsize{\textbf{DMPlug (ours)}} 
&\scriptsize{\textbf{0.067}}
&\scriptsize{\textbf{31.25}}
&\scriptsize{\textbf{0.878}}

&\scriptsize{\underbar{0.079}}
&\scriptsize{\textbf{30.25}}
&\scriptsize{\textbf{0.871}}

&\scriptsize{\textbf{0.039}}
&\scriptsize{\textbf{34.03}}
&\scriptsize{\textbf{0.936}}

&\scriptsize{\textbf{0.038}}
&\scriptsize{\textbf{33.01}}
&\scriptsize{\textbf{0.931}}
\\
\scriptsize{\textbf{Ours vs. Best compe.}}
&\scriptsize{\textcolor{darkpastelgreen}{$-0.003$}}
&\scriptsize{\textcolor{darkpastelgreen}{$+2.36$}}
&\scriptsize{\textcolor{darkpastelgreen}{$+0.030$}}

&\scriptsize{\textcolor{red}{$+0.003$}}
&\scriptsize{\textcolor{darkpastelgreen}{$+2.22$}}
&\scriptsize{\textcolor{darkpastelgreen}{$+0.028$}}

&\scriptsize{\textcolor{darkpastelgreen}{$-0.000$}}
&\scriptsize{\textcolor{darkpastelgreen}{$+1.79$}}
&\scriptsize{\textcolor{darkpastelgreen}{$+0.012$}}

&\scriptsize{\textcolor{darkpastelgreen}{$-0.006$}}
&\scriptsize{\textcolor{darkpastelgreen}{$+2.06$}}
&\scriptsize{\textcolor{darkpastelgreen}{$+0.018$}}
\\

\hline

\hline
\end{tabular}
}
\end{center}
\end{table}

\begin{table}[!htpb]
\caption{(\textcolor{red}{Linear IPs}) \textbf{Super-resolution} and \textbf{inpainting} on LSUN-bedroom~\cite{yu_lsun_2016} with additive Gaussian noise ($\sigma = 0.01$). (\textbf{Bold}: best, \underbar{under}: second best, \textcolor{darkpastelgreen}{green}: performance increase, \textcolor{red}{red}: performance decrease)}
\label{tab:sr_inp_bedroom}
\begin{center}
\setlength{\tabcolsep}{1.0mm}{
\begin{tabular}{c c c c c c c}
\hline

&\multicolumn{3}{c}{\scriptsize{\textbf{Super-resolution ($4 \times$)}}}
&\multicolumn{3}{c}{\scriptsize{\textbf{Inpainting (Random $70\%$)}}}

\\

\cmidrule(lr){2-4}
\cmidrule(lr){5-7}

&\scriptsize{LPIPS$\downarrow$}
&\scriptsize{PSNR$\uparrow$}
&\scriptsize{SSIM$\uparrow$}

&\scriptsize{LPIPS$\downarrow$}
&\scriptsize{PSNR$\uparrow$}
&\scriptsize{SSIM$\uparrow$}
\\
\hline

\scriptsize{ADMM-PnP~\cite{chan_plug-and-play_2016}}
&\scriptsize{0.358}
&\scriptsize{24.09}
&\scriptsize{0.728}

&\scriptsize{0.223}
&\scriptsize{27.75}
&\scriptsize{0.842}
\\

\scriptsize{DMPS~\cite{meng_diffusion_2024}}
&\scriptsize{0.172}
&\scriptsize{\underbar{25.54}}
&\scriptsize{\underbar{0.772}}

&\scriptsize{0.320}
&\scriptsize{23.09}
&\scriptsize{0.712}
\\

\scriptsize{DDRM~\cite{kawar_denoising_2022}}
&\scriptsize{0.415}
&\scriptsize{22.73}
&\scriptsize{0.644}

&\scriptsize{0.273}
&\scriptsize{23.66}
&\scriptsize{0.673}
\\

\scriptsize{MCG~\cite{chung_improving_2022}}
&\scriptsize{1.069}
&\scriptsize{15.00}
&\scriptsize{0.176}

&\scriptsize{1.271}
&\scriptsize{11.15}
&\scriptsize{0.055}
\\

\scriptsize{ILVR~\cite{choi_ilvr_2021}}
&\scriptsize{0.476}
&\scriptsize{18.89}
&\scriptsize{0.445}

&\scriptsize{0.581}
&\scriptsize{14.79}
&\scriptsize{0.407}
\\

\scriptsize{DPS~\cite{chung_diffusion_2023}} 

&\scriptsize{0.195}
&\scriptsize{25.10}
&\scriptsize{0.737}

&\scriptsize{0.094}
&\scriptsize{\underbar{28.95}}
&\scriptsize{\underbar{0.868}}
\\

\scriptsize{ReSample~\cite{song_solving_2023}}

&\scriptsize{\underbar{0.137}}
&\scriptsize{24.72}
&\scriptsize{0.762}

&\scriptsize{\underbar{0.069}}
&\scriptsize{26.43}
&\scriptsize{0.861}
\\

\scriptsize{\textbf{DMPlug (ours)}}

&\scriptsize{\textbf{0.136}}
&\scriptsize{\textbf{26.93}}
&\scriptsize{\textbf{0.801}}

&\scriptsize{\textbf{0.064}}
&\scriptsize{\textbf{30.15}}
&\scriptsize{\textbf{0.905}}
\\
\scriptsize{\textbf{Ours vs. Best compe.}}
&\scriptsize{\textcolor{darkpastelgreen}{$-0.001$}}
&\scriptsize{\textcolor{darkpastelgreen}{$+1.39$}}
&\scriptsize{\textcolor{darkpastelgreen}{$+0.029$}}

&\scriptsize{\textcolor{darkpastelgreen}{$-0.005$}}
&\scriptsize{\textcolor{darkpastelgreen}{$+1.20$}}
&\scriptsize{\textcolor{darkpastelgreen}{$+0.037$}}

\\

\hline

\hline
\end{tabular}
}
\end{center}
\end{table}

\begin{table}[!htpb]
\caption{(\textcolor{red}{Linear IPs}) \textbf{Super-resolution} and \textbf{inpainting} on CelebA~\cite{liu_deep_2015} with additive Gaussian noise ($\sigma = 0.01$). (\textbf{Bold}: best, \underbar{under}: second best, \textcolor{darkpastelgreen}{green}: performance increase, \textcolor{red}{red}: performance decrease)}
\label{tab:sr_inp_more}
\begin{center}
\setlength{\tabcolsep}{1.0mm}{
\begin{tabular}{c c c c c c c}
\hline

&\multicolumn{3}{c}{\scriptsize{\textbf{Super-resolution ($4 \times$)}}}
&\multicolumn{3}{c}{\scriptsize{\textbf{Inpainting (Random $70\%$)}}}

\\

\cmidrule(lr){2-4}
\cmidrule(lr){5-7}

&\scriptsize{LPIPS$\downarrow$}
&\scriptsize{PSNR$\uparrow$}
&\scriptsize{SSIM$\uparrow$}

&\scriptsize{LPIPS$\downarrow$}
&\scriptsize{PSNR$\uparrow$}
&\scriptsize{SSIM$\uparrow$}
\\
\hline

\scriptsize{FPS~\cite{dou_diffusion_2023}}
&\scriptsize{\underbar{0.149}}
&\scriptsize{29.12}
&\scriptsize{\underbar{0.858}}

&\scriptsize{\underbar{0.064}}
&\scriptsize{\underbar{32.06}}
&\scriptsize{\underbar{0.924}}
\\

\scriptsize{DiffPIR~\cite{zhu_denoising_2023}}
&\scriptsize{0.203}
&\scriptsize{\textbf{31.55}}
&\scriptsize{0.857}

&\scriptsize{0.219}
&\scriptsize{31.22}
&\scriptsize{0.866}
\\

\scriptsize{DDNM~\cite{wang_zero-shot_2022}}
&\scriptsize{0.193}
&\scriptsize{29.21}
&\scriptsize{0.836}

&\scriptsize{0.224}
&\scriptsize{27.89}
&\scriptsize{0.799}
\\

\scriptsize{PSLD~\cite{rout_solving_2023}}
&\scriptsize{0.243}
&\scriptsize{26.45}
&\scriptsize{0.682}

&\scriptsize{0.213}
&\scriptsize{27.65}
&\scriptsize{0.785}
\\

\scriptsize{DMPlug (ours)}
&\scriptsize{\textbf{0.067}}
&\scriptsize{\underbar{31.25}}
&\scriptsize{\textbf{0.878}}

&\scriptsize{\textbf{0.039}}
&\scriptsize{\textbf{34.03}}
&\scriptsize{\textbf{0.936}}

\\
\scriptsize{\textbf{Ours vs. Best compe.}}
&\scriptsize{\textcolor{darkpastelgreen}{$-0.082$}}
&\scriptsize{\textcolor{red}{$-0.30$}}
&\scriptsize{\textcolor{darkpastelgreen}{$+0.020$}}

&\scriptsize{\textcolor{darkpastelgreen}{$-0.025$}}
&\scriptsize{\textcolor{darkpastelgreen}{$+1.97$}}
&\scriptsize{\textcolor{darkpastelgreen}{$+0.012$}}

\\

\hline

\hline
\end{tabular}
}
\end{center}
\end{table}

\subsection{Qualitative results}
\label{app:qual_res}

\begin{figure}[!htbp]
    \centering 
    \vspace{-1em}
    \includegraphics[width=1\linewidth]{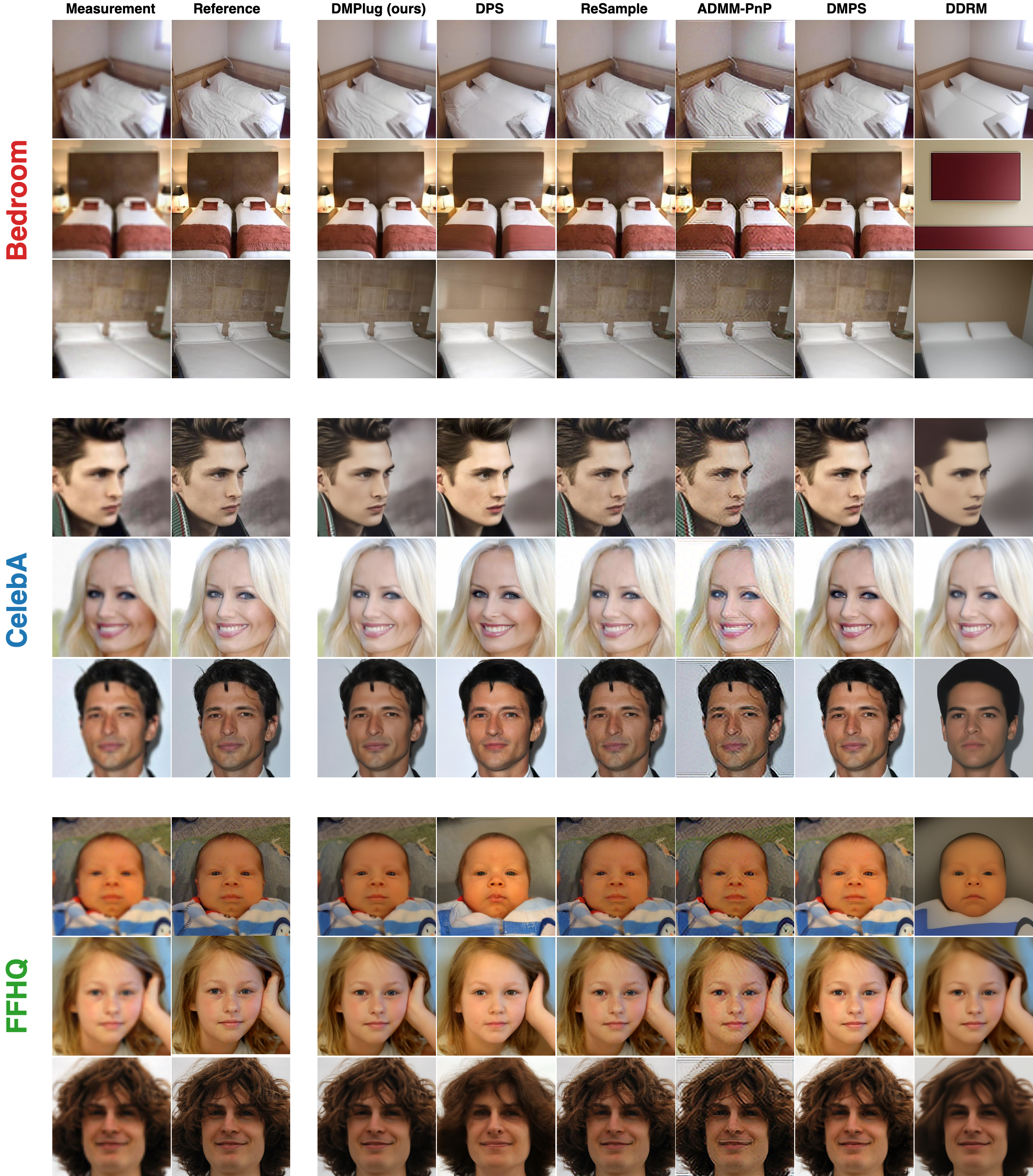}
    \vspace{-1em}
    \caption{(\textcolor{red}{Linear IP}) Visualization of sample results from the plug-in method (\textbf{Ours}) and competing methods for \textbf{$4 \times$ super-solution}. All measurements contain Gaussian noise with $\sigma = 0.01$.}
    \label{fig:ap_sr}
    \vspace{-1em}
\end{figure}

\begin{figure}[!htbp]
    \centering 
    \vspace{-1em}
    \includegraphics[width=1\linewidth]{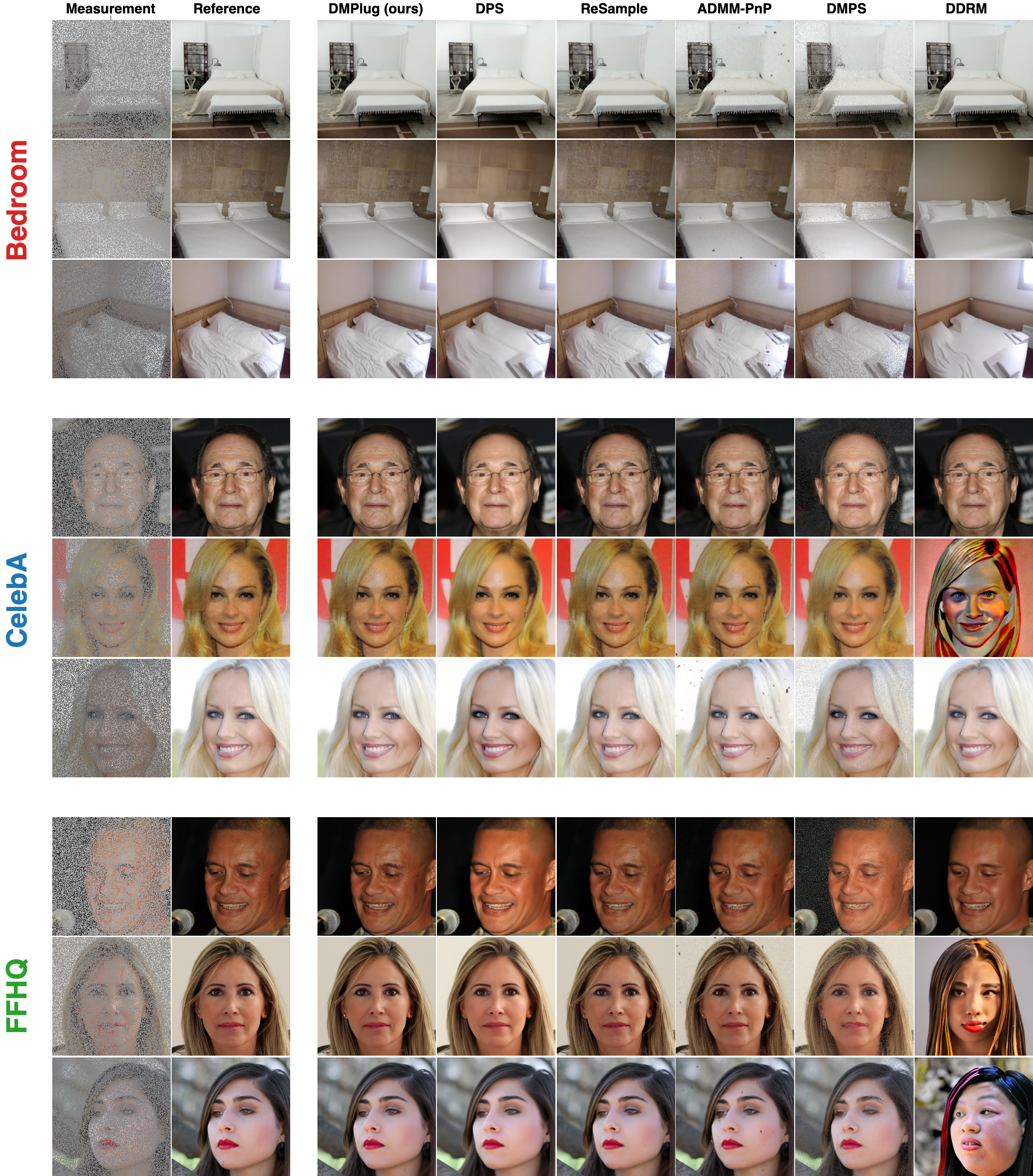}
    \vspace{-1em}
    \caption{(\textcolor{red}{Linear IP}) Visualization of sample results from the plug-in method (\textbf{Ours}) and competing methods for \textbf{inpainting (random $70\%$)}. All measurements contain Gaussian noise with $\sigma = 0.01$.}
    \label{fig:ap_inp}
    \vspace{-1em}
\end{figure}

\begin{figure}[!htbp]
    \centering 
    \vspace{-1em}
    \includegraphics[width=1\linewidth]{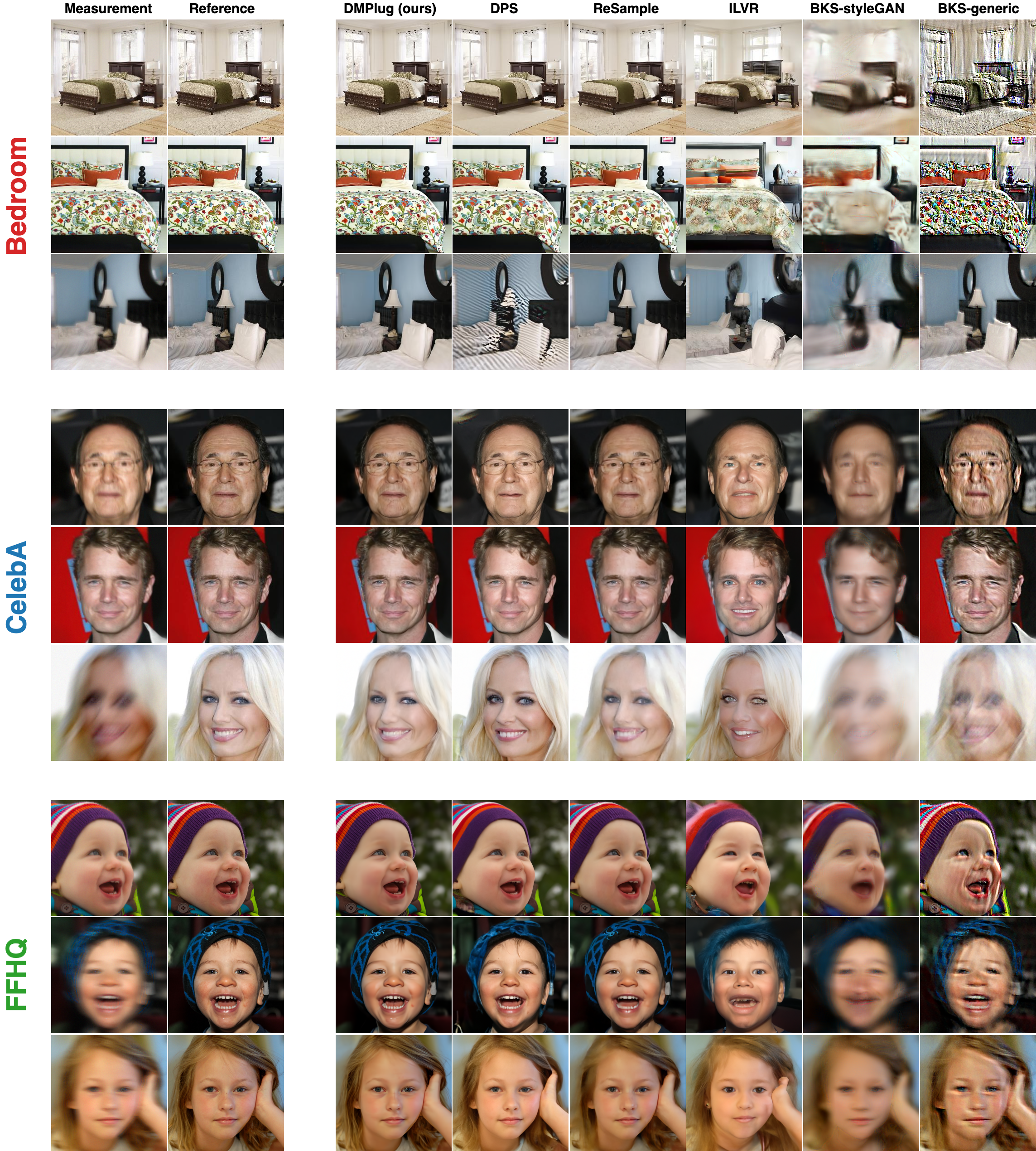}
    \vspace{-1em}
    \caption{(\textcolor{red}{Nonlinear IP}) Visualization of sample results from the plug-in method (\textbf{Ours}) and competing methods for \textbf{nonlinear deblurring}. All measurements contain Gaussian noise with $\sigma = 0.01$.}
    \label{fig:ap_nonuniform}
    \vspace{-1em}
\end{figure}

\begin{figure}[!htbp]
    \centering 
    \vspace{-1em}
    \includegraphics[width=1\linewidth]{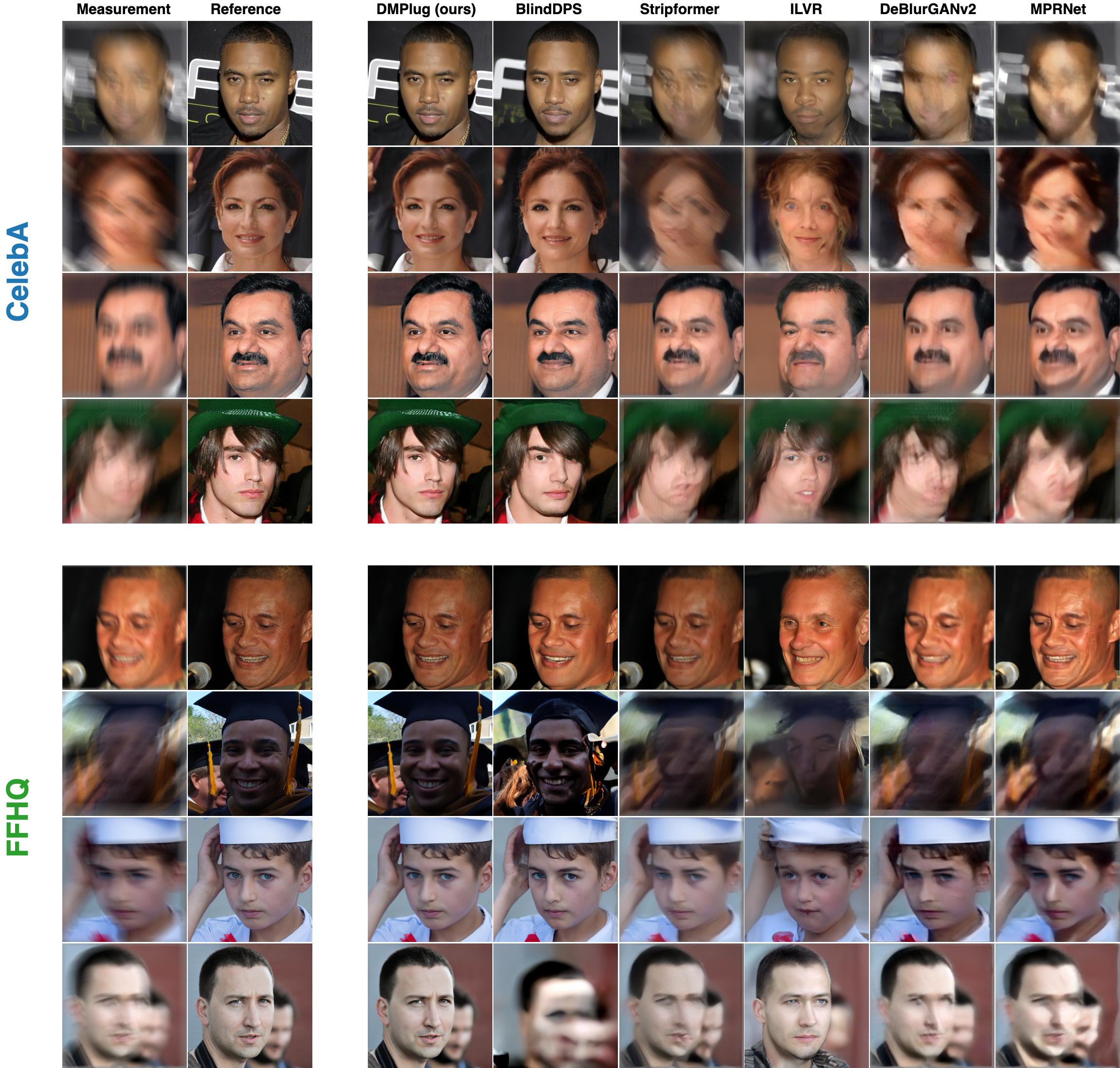}
    \vspace{-1em}
    \caption{(\textcolor{red}{Nonlinear IP}) Visualization of sample results from the plug-in method (\textbf{Ours}) and competing methods for \textbf{BID (motion)}. All measurements contain Gaussian noise with $\sigma = 0.01$.}
    \label{fig:ap_bid_motion}
    \vspace{-1em}
\end{figure}

\begin{figure}[!htbp]
    \centering 
    \vspace{-1em}
    \includegraphics[width=1\linewidth]{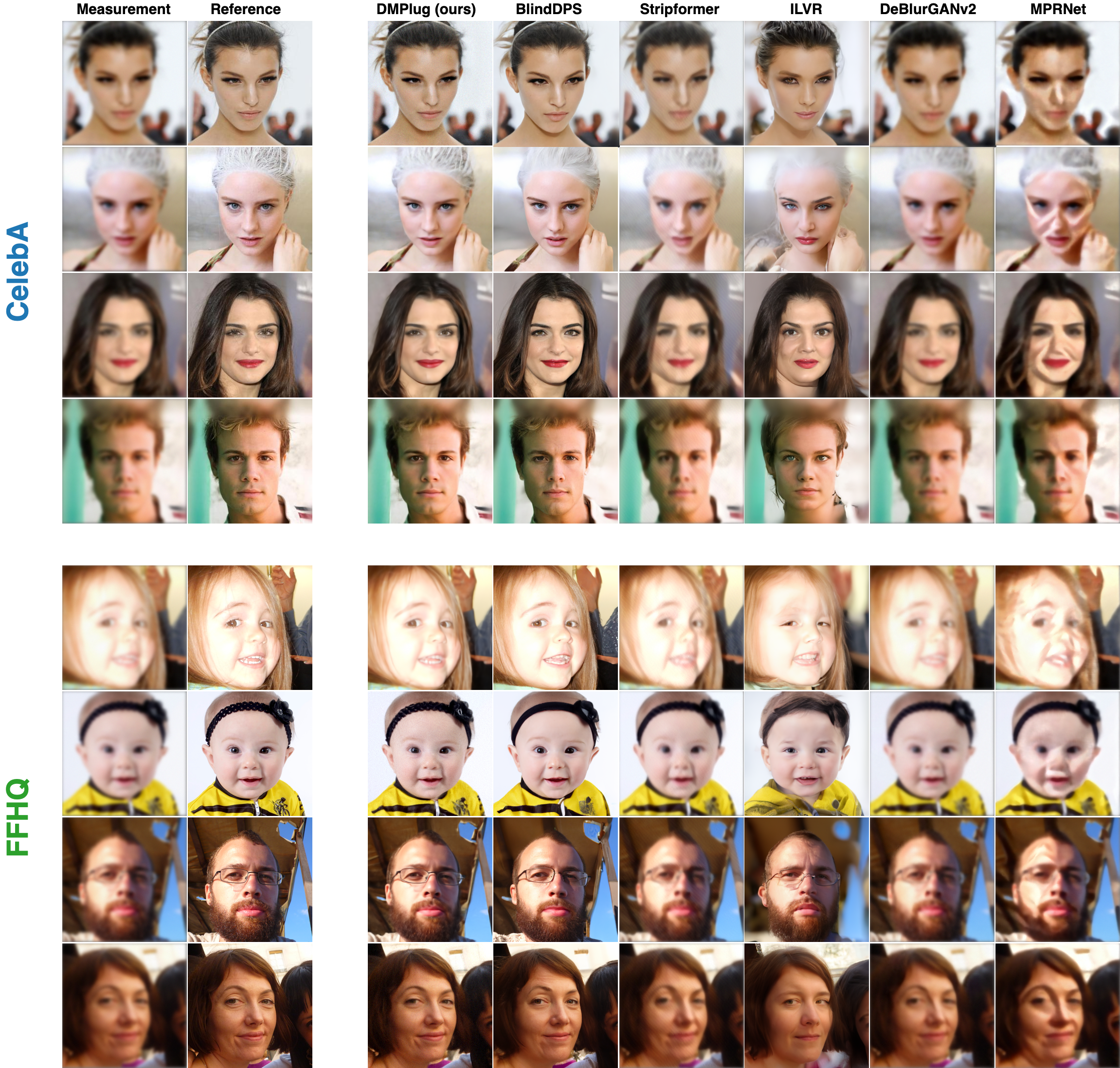}
    \vspace{-1em}
    \caption{(\textcolor{red}{Nonlinear IP}) Visualization of sample results from the plug-in method (\textbf{Ours}) and competing methods for \textbf{BID (Gaussian)}. All measurements contain Gaussian noise with $\sigma = 0.01$.}
    \label{fig:ap_bid_gaussian}
    \vspace{-1em}
\end{figure}

\begin{figure}[!htbp]
    \centering 
    \vspace{-1em}
    \includegraphics[width=1\linewidth]{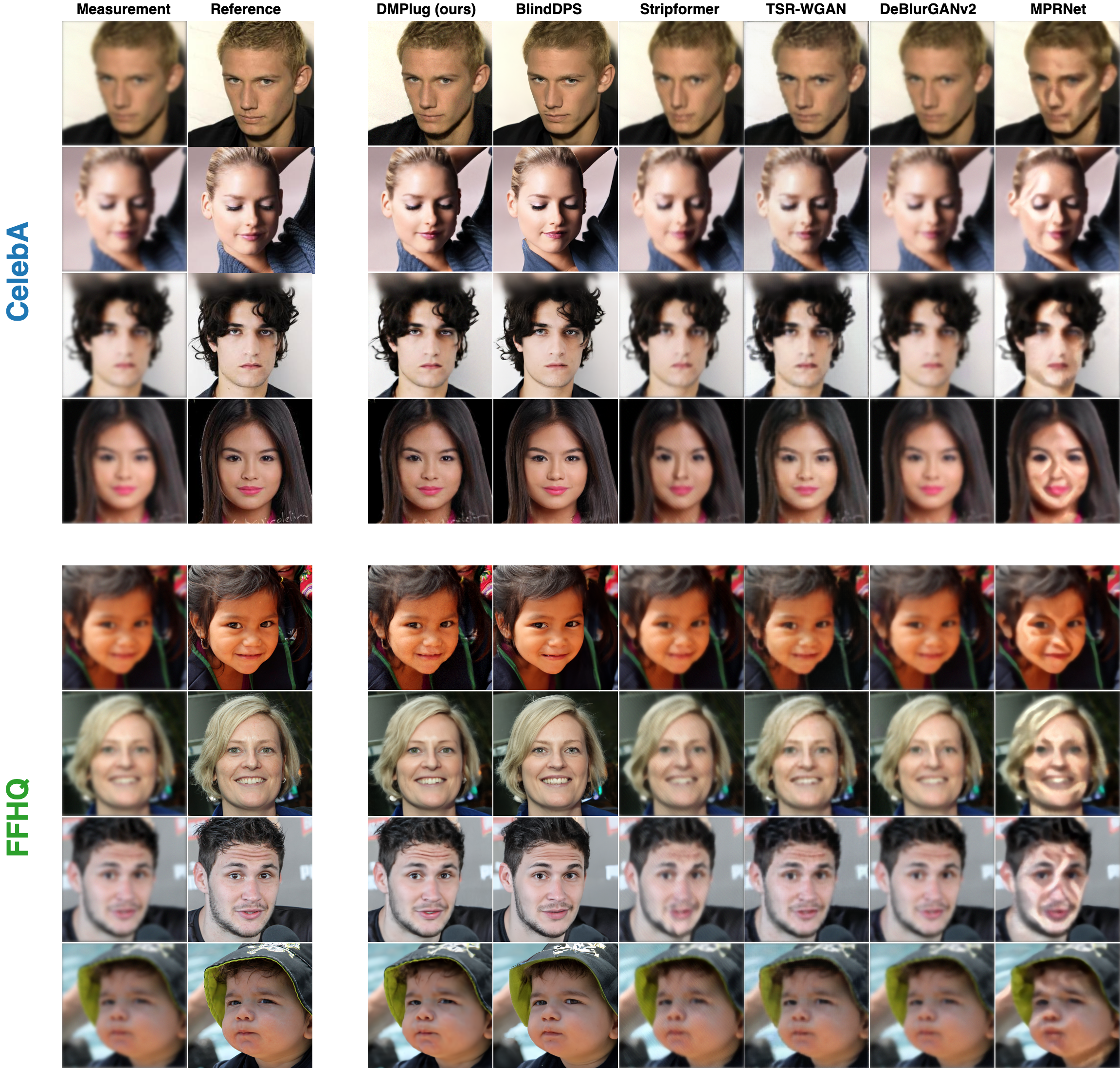}
    \vspace{-1em}
    \caption{(\textcolor{red}{Nonlinear IP}) Visualization of sample results from the plug-in method (\textbf{Ours}) and competing methods for \textbf{BID with turbulence}. All measurements contain Gaussian noise with $\sigma = 0.01$.}
    \label{fig:ap_tur}
    \vspace{-1em}
\end{figure}

\begin{figure}[!htbp]
    \centering 
    \vspace{-1em}
    \includegraphics[width=1\linewidth]{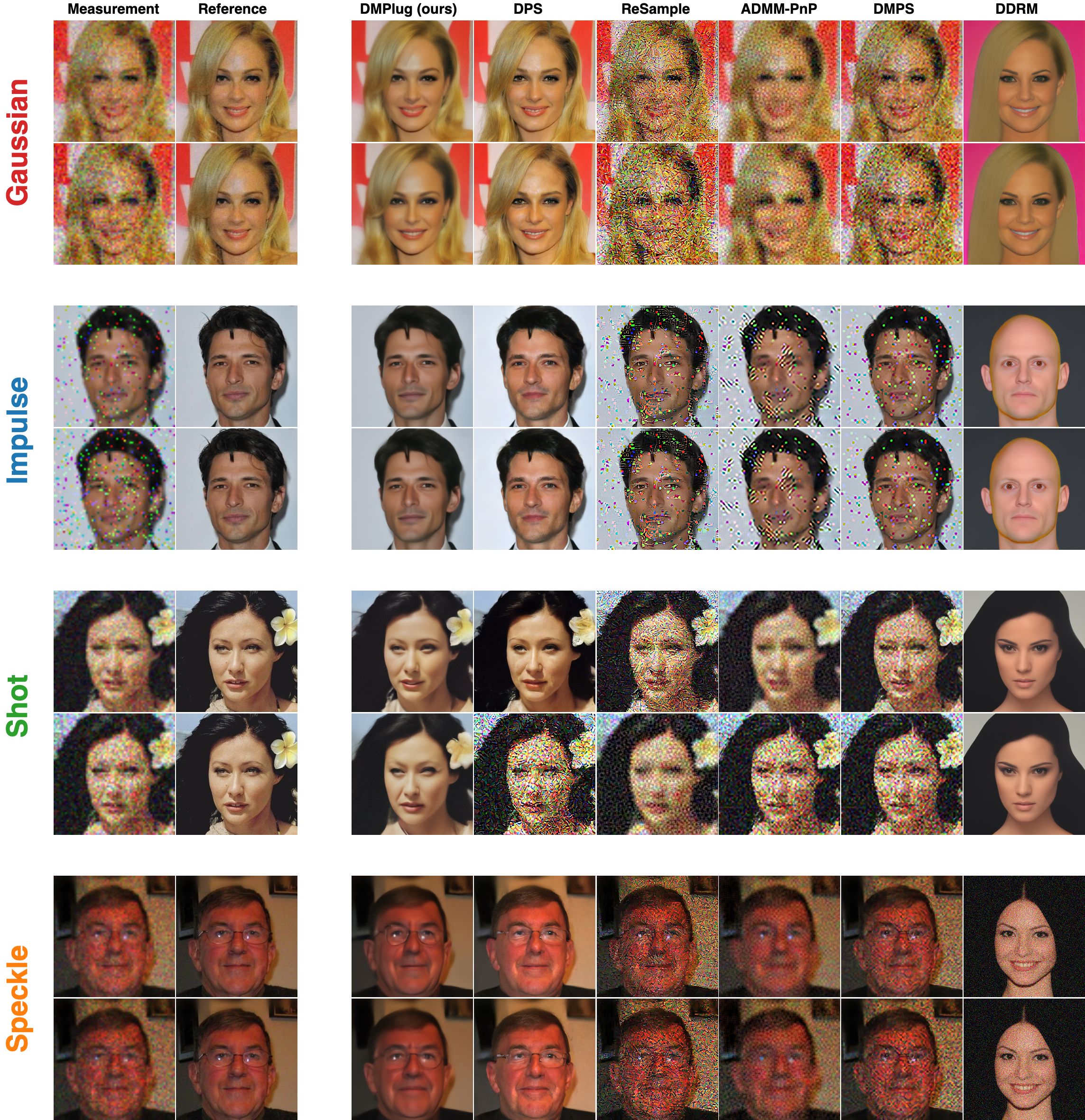}
    \vspace{-1em}
    \caption{(\textcolor{red}{Robustness}) Visualization of sample results from the plug-in method (\textbf{Ours}) and competing methods for \textbf{$4 \times$ super-resolution}. We generate measurements with four types of noise—Gaussian, impulse, shot, and speckle noise—across two different noise levels: low (level-1) and high (level-2), following~\cite{hendrycks_benchmarking_2019}. (top: low-level noise; bottom: high-level noise)}
    \label{fig:ap_robust_sr}
    \vspace{-1em}
\end{figure}

\begin{figure}[!htbp]
    \centering 
    \vspace{-1em}
    \includegraphics[width=1\linewidth]{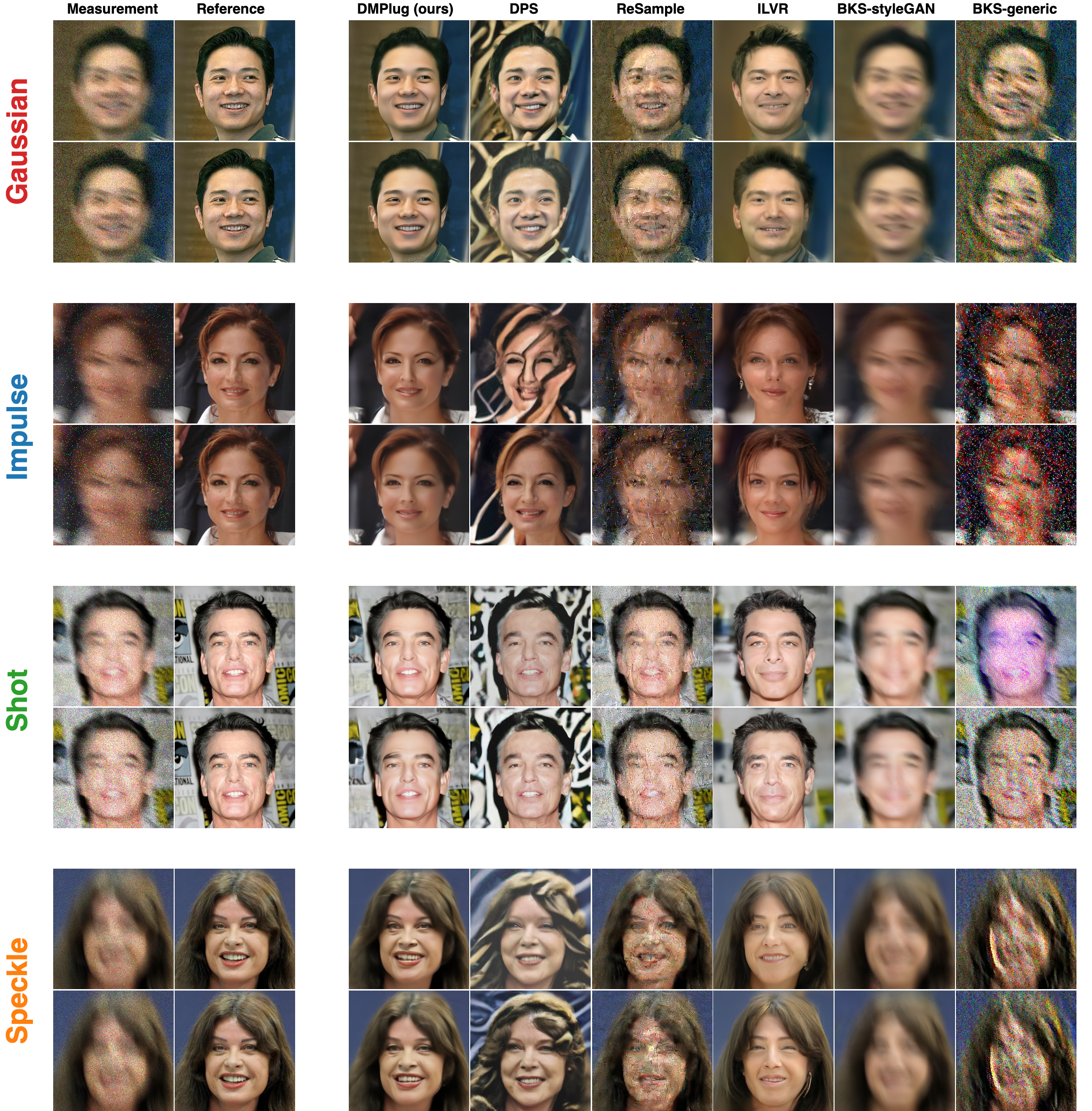}
    \vspace{-1em}
    \caption{(\textcolor{red}{Robustness}) Visualization of sample results from the plug-in method (\textbf{Ours}) and competing methods for \textbf{nonlinear deblurring}. We generate measurements with four types of noise—Gaussian, impulse, shot, and speckle noise—across two different noise levels: low (level-1) and high (level-2), following~\cite{hendrycks_benchmarking_2019}. (top: low-level noise; bottom: high-level noise)}
    \label{fig:ap_robust_blur}
    \vspace{-1em}
\end{figure}

\section{More early stopping results}
\label{app:es_res}

\begin{figure}[!htbp]
    \centering 
    \vspace{-1em}
    \includegraphics[width=1\linewidth]{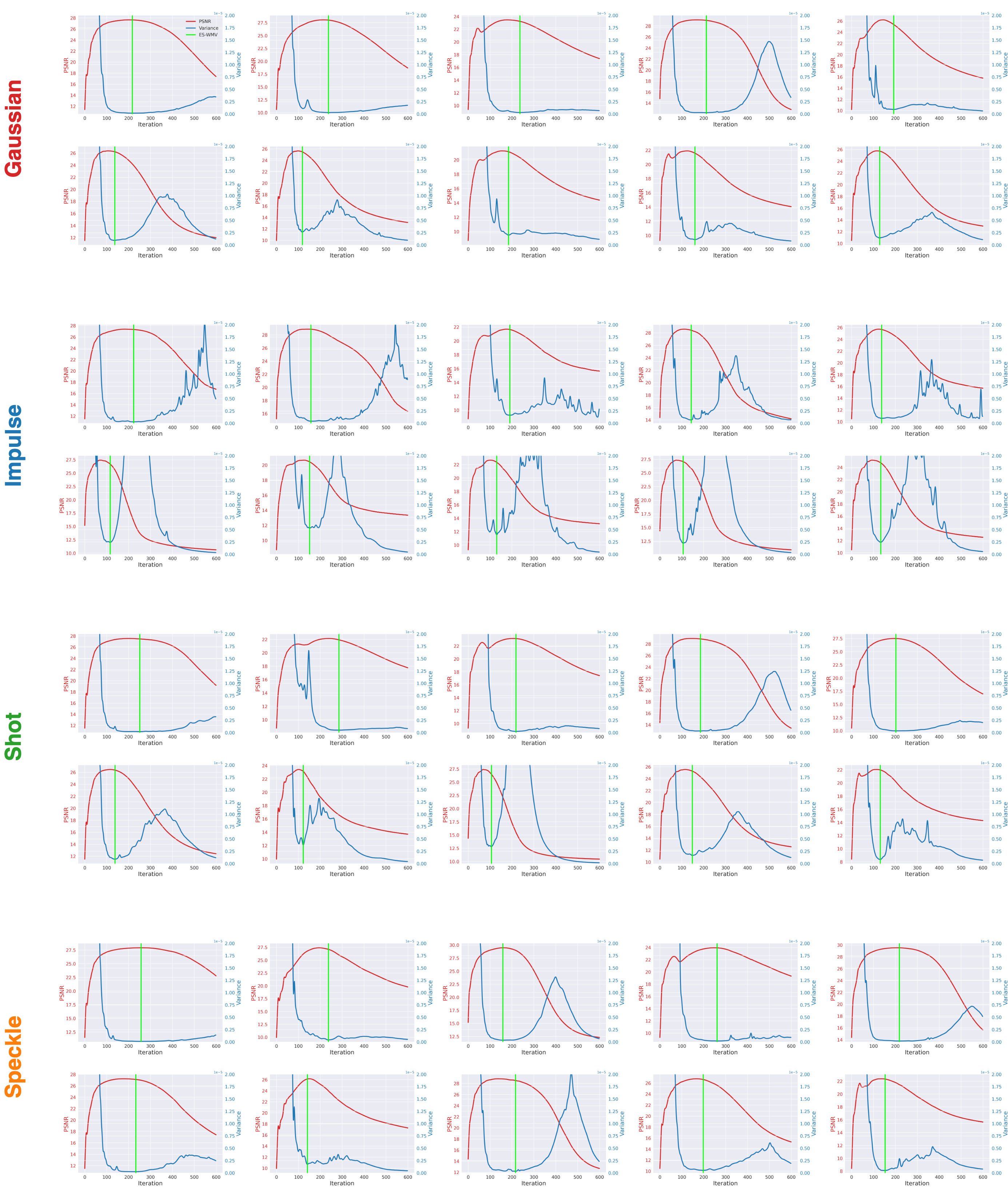}
    \vspace{-1em}
    \caption{(\textcolor{red}{Early stopping}) Our DMPlgu with ES-WMV~\cite{wang_early_2023} for \textbf{$4 \times$ super-resolution} with different types and levels of noise. (top: low-level noise; bottom: high-level noise). \textcolor{myred}{Red curves} are PSNR curves, and \textcolor{myblue}{blue curves} are VAR curves. The \textcolor{mygreen}{green bars} indicate the detected ES point.}
    \label{fig:ap_es_sr}
    \vspace{-1em}
\end{figure}

\begin{figure}[!htbp]
    \centering 
    \vspace{-1em}
    \includegraphics[width=1\linewidth]{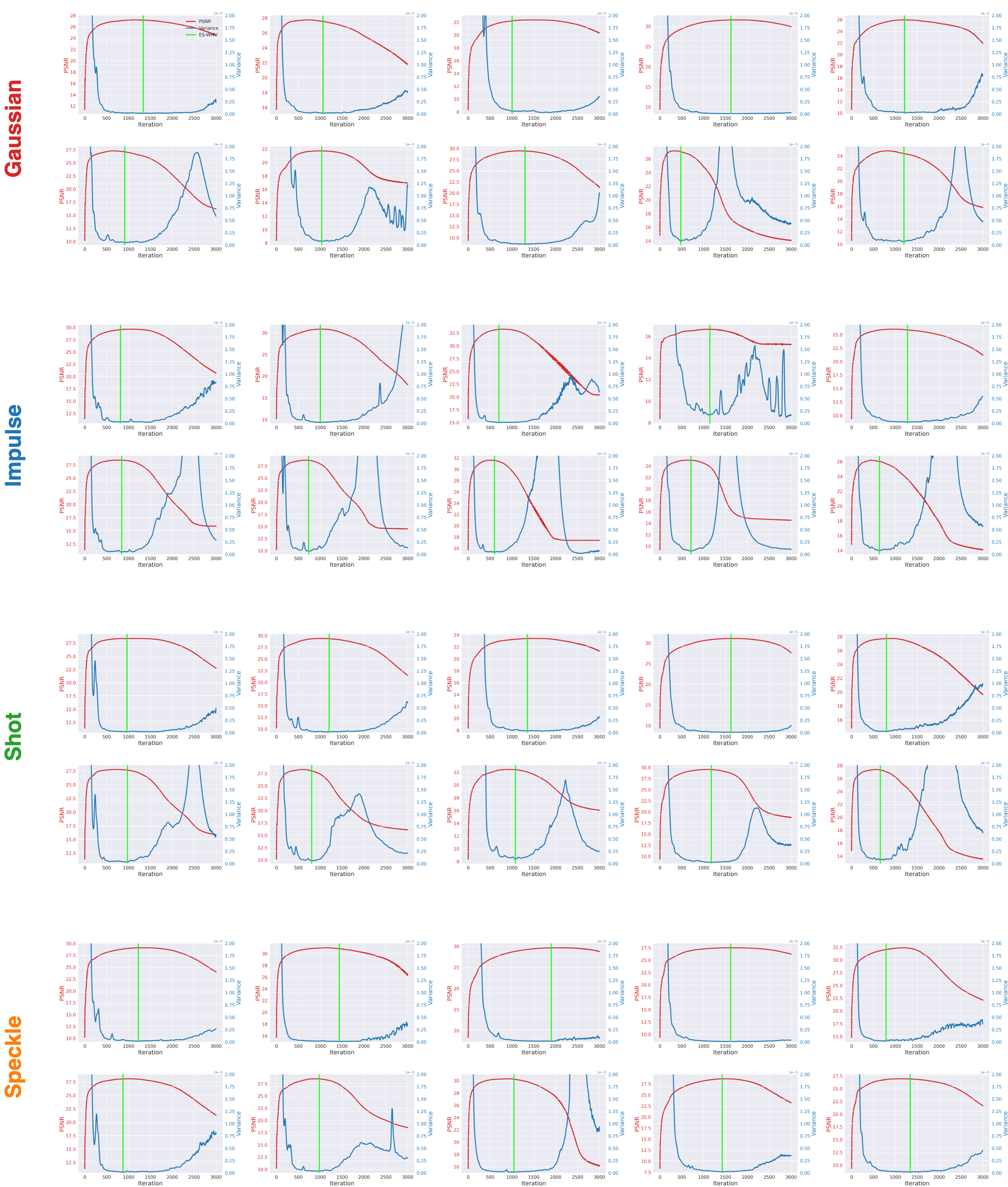}
    \vspace{-1em}
    \caption{(\textcolor{red}{Early stopping}) Our DMPlug with ES-WMV~\cite{wang_early_2023} for \textbf{nonlinear deblurring} with different types and levels of noise. (top: low-level noise; bottom: high-level noise). \textcolor{myred}{Red curves} are PSNR curves, and \textcolor{myblue}{blue curves} are VAR curves. The \textcolor{mygreen}{green bars} indicate the detected ES point.}
    \label{fig:ap_es_deblur}
    \vspace{-1em}
\end{figure}


\clearpage 

\section*{NeurIPS Paper Checklist}

\begin{enumerate}

\item {\bf Claims}
    \item[] Question: Do the main claims made in the abstract and introduction accurately reflect the paper's contributions and scope?
    \item[] Answer: \answerYes{} 
    \item[] Justification: We clearly claim the scope and contributions of this paper in both abstract and introduction.
    \item[] Guidelines:
    \begin{itemize}
        \item The answer NA means that the abstract and introduction do not include the claims made in the paper.
        \item The abstract and/or introduction should clearly state the claims made, including the contributions made in the paper and important assumptions and limitations. A No or NA answer to this question will not be perceived well by the reviewers. 
        \item The claims made should match theoretical and experimental results, and reflect how much the results can be expected to generalize to other settings. 
        \item It is fine to include aspirational goals as motivation as long as it is clear that these goals are not attained by the paper. 
    \end{itemize}

\item {\bf Limitations}
    \item[] Question: Does the paper discuss the limitations of the work performed by the authors?
    \item[] Answer: \answerYes{}
    \item[] Justification: We discuss the limitations of this work in \cref{sec:dis}.
    \item[] Guidelines:
    \begin{itemize}
        \item The answer NA means that the paper has no limitation while the answer No means that the paper has limitations, but those are not discussed in the paper. 
        \item The authors are encouraged to create a separate "Limitations" section in their paper.
        \item The paper should point out any strong assumptions and how robust the results are to violations of these assumptions (e.g., independence assumptions, noiseless settings, model well-specification, asymptotic approximations only holding locally). The authors should reflect on how these assumptions might be violated in practice and what the implications would be.
        \item The authors should reflect on the scope of the claims made, e.g., if the approach was only tested on a few datasets or with a few runs. In general, empirical results often depend on implicit assumptions, which should be articulated.
        \item The authors should reflect on the factors that influence the performance of the approach. For example, a facial recognition algorithm may perform poorly when image resolution is low or images are taken in low lighting. Or a speech-to-text system might not be used reliably to provide closed captions for online lectures because it fails to handle technical jargon.
        \item The authors should discuss the computational efficiency of the proposed algorithms and how they scale with dataset size.
        \item If applicable, the authors should discuss possible limitations of their approach to address problems of privacy and fairness.
        \item While the authors might fear that complete honesty about limitations might be used by reviewers as grounds for rejection, a worse outcome might be that reviewers discover limitations that aren't acknowledged in the paper. The authors should use their best judgment and recognize that individual actions in favor of transparency play an important role in developing norms that preserve the integrity of the community. Reviewers will be specifically instructed to not penalize honesty concerning limitations.
    \end{itemize}

\item {\bf Theory Assumptions and Proofs}
    \item[] Question: For each theoretical result, does the paper provide the full set of assumptions and a complete (and correct) proof?
    \item[] Answer: \answerYes{} 
    \item[] Justification: All the formulas are numbered and cross-referenced.
    \item[] Guidelines:
    \begin{itemize}
        \item The answer NA means that the paper does not include theoretical results. 
        \item All the theorems, formulas, and proofs in the paper should be numbered and cross-referenced.
        \item All assumptions should be clearly stated or referenced in the statement of any theorems.
        \item The proofs can either appear in the main paper or the supplemental material, but if they appear in the supplemental material, the authors are encouraged to provide a short proof sketch to provide intuition. 
        \item Inversely, any informal proof provided in the core of the paper should be complemented by formal proofs provided in appendix or supplemental material.
        \item Theorems and Lemmas that the proof relies upon should be properly referenced. 
    \end{itemize}

    \item {\bf Experimental Result Reproducibility}
    \item[] Question: Does the paper fully disclose all the information needed to reproduce the main experimental results of the paper to the extent that it affects the main claims and/or conclusions of the paper (regardless of whether the code and data are provided or not)?
    \item[] Answer: \answerYes{} 
    \item[] Justification: We provide all the experiment details in \cref{sec:exps}, \cref{app:setup} and \cref{app:imp}.
    \item[] Guidelines:
    \begin{itemize}
        \item The answer NA means that the paper does not include experiments.
        \item If the paper includes experiments, a No answer to this question will not be perceived well by the reviewers: Making the paper reproducible is important, regardless of whether the code and data are provided or not.
        \item If the contribution is a dataset and/or model, the authors should describe the steps taken to make their results reproducible or verifiable. 
        \item Depending on the contribution, reproducibility can be accomplished in various ways. For example, if the contribution is a novel architecture, describing the architecture fully might suffice, or if the contribution is a specific model and empirical evaluation, it may be necessary to either make it possible for others to replicate the model with the same dataset, or provide access to the model. In general. releasing code and data is often one good way to accomplish this, but reproducibility can also be provided via detailed instructions for how to replicate the results, access to a hosted model (e.g., in the case of a large language model), releasing of a model checkpoint, or other means that are appropriate to the research performed.
        \item While NeurIPS does not require releasing code, the conference does require all submissions to provide some reasonable avenue for reproducibility, which may depend on the nature of the contribution. For example
        \begin{enumerate}
            \item If the contribution is primarily a new algorithm, the paper should make it clear how to reproduce that algorithm.
            \item If the contribution is primarily a new model architecture, the paper should describe the architecture clearly and fully.
            \item If the contribution is a new model (e.g., a large language model), then there should either be a way to access this model for reproducing the results or a way to reproduce the model (e.g., with an open-source dataset or instructions for how to construct the dataset).
            \item We recognize that reproducibility may be tricky in some cases, in which case authors are welcome to describe the particular way they provide for reproducibility. In the case of closed-source models, it may be that access to the model is limited in some way (e.g., to registered users), but it should be possible for other researchers to have some path to reproducing or verifying the results.
        \end{enumerate}
    \end{itemize}

\item {\bf Open access to data and code}
    \item[] Question: Does the paper provide open access to the data and code, with sufficient instructions to faithfully reproduce the main experimental results, as described in supplemental material?
    \item[] Answer: \answerYes{} 
    \item[] Justification: We provide all the essential code for this paper.
    \item[] Guidelines:
    \begin{itemize}
        \item The answer NA means that paper does not include experiments requiring code.
        \item Please see the NeurIPS code and data submission guidelines (\url{https://nips.cc/public/guides/CodeSubmissionPolicy}) for more details.
        \item While we encourage the release of code and data, we understand that this might not be possible, so “No” is an acceptable answer. Papers cannot be rejected simply for not including code, unless this is central to the contribution (e.g., for a new open-source benchmark).
        \item The instructions should contain the exact command and environment needed to run to reproduce the results. See the NeurIPS code and data submission guidelines (\url{https://nips.cc/public/guides/CodeSubmissionPolicy}) for more details.
        \item The authors should provide instructions on data access and preparation, including how to access the raw data, preprocessed data, intermediate data, and generated data, etc.
        \item The authors should provide scripts to reproduce all experimental results for the new proposed method and baselines. If only a subset of experiments are reproducible, they should state which ones are omitted from the script and why.
        \item At submission time, to preserve anonymity, the authors should release anonymized versions (if applicable).
        \item Providing as much information as possible in supplemental material (appended to the paper) is recommended, but including URLs to data and code is permitted.
    \end{itemize}

\item {\bf Experimental Setting/Details}
    \item[] Question: Does the paper specify all the training and test details (e.g., data splits, hyperparameters, how they were chosen, type of optimizer, etc.) necessary to understand the results?
    \item[] Answer: \answerYes{} 
    \item[] Justification: We provide all the experiment details in \cref{sec:exps}, \cref{app:setup} and \cref{app:imp}.
    \item[] Guidelines:
    \begin{itemize}
        \item The answer NA means that the paper does not include experiments.
        \item The experimental setting should be presented in the core of the paper to a level of detail that is necessary to appreciate the results and make sense of them.
        \item The full details can be provided either with the code, in appendix, or as supplemental material.
    \end{itemize}

\item {\bf Experiment Statistical Significance}
    \item[] Question: Does the paper report error bars suitably and correctly defined or other appropriate information about the statistical significance of the experiments?
    \item[] Answer: \answerNo{}
    \item[] Justification: First, we conduct extensive experiments in this paper and report the mean results. We do not anticipate significant fluctuations in the results. Second, we are afraid that adding the statistical significance of the experiments will mess this paper up because our tables are already very dense, but we are willing to provide them if needed.
    \item[] Guidelines:
    \begin{itemize}
        \item The answer NA means that the paper does not include experiments.
        \item The authors should answer "Yes" if the results are accompanied by error bars, confidence intervals, or statistical significance tests, at least for the experiments that support the main claims of the paper.
        \item The factors of variability that the error bars are capturing should be clearly stated (for example, train/test split, initialization, random drawing of some parameter, or overall run with given experimental conditions).
        \item The method for calculating the error bars should be explained (closed form formula, call to a library function, bootstrap, etc.)
        \item The assumptions made should be given (e.g., Normally distributed errors).
        \item It should be clear whether the error bar is the standard deviation or the standard error of the mean.
        \item It is OK to report 1-sigma error bars, but one should state it. The authors should preferably report a 2-sigma error bar than state that they have a 96\% CI, if the hypothesis of Normality of errors is not verified.
        \item For asymmetric distributions, the authors should be careful not to show in tables or figures symmetric error bars that would yield results that are out of range (e.g. negative error rates).
        \item If error bars are reported in tables or plots, The authors should explain in the text how they were calculated and reference the corresponding figures or tables in the text.
    \end{itemize}

\item {\bf Experiments Compute Resources}
    \item[] Question: For each experiment, does the paper provide sufficient information on the computer resources (type of compute workers, memory, time of execution) needed to reproduce the experiments?
    \item[] Answer: \answerYes{} 
    \item[] Justification: We provide the experiment compute resources in \cref{app:imp_ours}.
    \item[] Guidelines:
    \begin{itemize}
        \item The answer NA means that the paper does not include experiments.
        \item The paper should indicate the type of compute workers CPU or GPU, internal cluster, or cloud provider, including relevant memory and storage.
        \item The paper should provide the amount of compute required for each of the individual experimental runs as well as estimate the total compute. 
        \item The paper should disclose whether the full research project required more compute than the experiments reported in the paper (e.g., preliminary or failed experiments that didn't make it into the paper). 
    \end{itemize}
    
\item {\bf Code Of Ethics}
    \item[] Question: Does the research conducted in the paper conform, in every respect, with the NeurIPS Code of Ethics \url{https://neurips.cc/public/EthicsGuidelines}?
    \item[] Answer: \answerYes{} 
    \item[] Justification: This paper is strictly with the NeurIPS Code of Ethics.
    \item[] Guidelines:
    \begin{itemize}
        \item The answer NA means that the authors have not reviewed the NeurIPS Code of Ethics.
        \item If the authors answer No, they should explain the special circumstances that require a deviation from the Code of Ethics.
        \item The authors should make sure to preserve anonymity (e.g., if there is a special consideration due to laws or regulations in their jurisdiction).
    \end{itemize}

\item {\bf Broader Impacts}
    \item[] Question: Does the paper discuss both potential positive societal impacts and negative societal impacts of the work performed?
    \item[] Answer: \answerYes{} 
    \item[] Justification: We discuss the societal impacts in \cref{sec:dis}.
    \item[] Guidelines:
    \begin{itemize}
        \item The answer NA means that there is no societal impact of the work performed.
        \item If the authors answer NA or No, they should explain why their work has no societal impact or why the paper does not address societal impact.
        \item Examples of negative societal impacts include potential malicious or unintended uses (e.g., disinformation, generating fake profiles, surveillance), fairness considerations (e.g., deployment of technologies that could make decisions that unfairly impact specific groups), privacy considerations, and security considerations.
        \item The conference expects that many papers will be foundational research and not tied to particular applications, let alone deployments. However, if there is a direct path to any negative applications, the authors should point it out. For example, it is legitimate to point out that an improvement in the quality of generative models could be used to generate deepfakes for disinformation. On the other hand, it is not needed to point out that a generic algorithm for optimizing neural networks could enable people to train models that generate Deepfakes faster.
        \item The authors should consider possible harms that could arise when the technology is being used as intended and functioning correctly, harms that could arise when the technology is being used as intended but gives incorrect results, and harms following from (intentional or unintentional) misuse of the technology.
        \item If there are negative societal impacts, the authors could also discuss possible mitigation strategies (e.g., gated release of models, providing defenses in addition to attacks, mechanisms for monitoring misuse, mechanisms to monitor how a system learns from feedback over time, improving the efficiency and accessibility of ML).
    \end{itemize}
    
\item {\bf Safeguards}
    \item[] Question: Does the paper describe safeguards that have been put in place for responsible release of data or models that have a high risk for misuse (e.g., pretrained language models, image generators, or scraped datasets)?
    \item[] Answer: \answerNA{} 
    \item[] Justification: This paper only uses existing pretrained models for zero-shot tasks.
    \item[] Guidelines:
    \begin{itemize}
        \item The answer NA means that the paper poses no such risks.
        \item Released models that have a high risk for misuse or dual-use should be released with necessary safeguards to allow for controlled use of the model, for example by requiring that users adhere to usage guidelines or restrictions to access the model or implementing safety filters. 
        \item Datasets that have been scraped from the Internet could pose safety risks. The authors should describe how they avoided releasing unsafe images.
        \item We recognize that providing effective safeguards is challenging, and many papers do not require this, but we encourage authors to take this into account and make a best faith effort.
    \end{itemize}

\item {\bf Licenses for existing assets}
    \item[] Question: Are the creators or original owners of assets (e.g., code, data, models), used in the paper, properly credited and are the license and terms of use explicitly mentioned and properly respected?
    \item[] Answer: \answerYes{} 
    \item[] Justification: We provide all the code and models we have used in \cref{app:imp}.
    \item[] Guidelines:
    \begin{itemize}
        \item The answer NA means that the paper does not use existing assets.
        \item The authors should cite the original paper that produced the code package or dataset.
        \item The authors should state which version of the asset is used and, if possible, include a URL.
        \item The name of the license (e.g., CC-BY 4.0) should be included for each asset.
        \item For scraped data from a particular source (e.g., website), the copyright and terms of service of that source should be provided.
        \item If assets are released, the license, copyright information, and terms of use in the package should be provided. For popular datasets, \url{paperswithcode.com/datasets} has curated licenses for some datasets. Their licensing guide can help determine the license of a dataset.
        \item For existing datasets that are re-packaged, both the original license and the license of the derived asset (if it has changed) should be provided.
        \item If this information is not available online, the authors are encouraged to reach out to the asset's creators.
    \end{itemize}

\item {\bf New Assets}
    \item[] Question: Are new assets introduced in the paper well documented and is the documentation provided alongside the assets?
    \item[] Answer: \answerYes{} 
    \item[] Justification: The new assets in this paper are well documented.
    \item[] Guidelines:
    \begin{itemize}
        \item The answer NA means that the paper does not release new assets.
        \item Researchers should communicate the details of the dataset/code/model as part of their submissions via structured templates. This includes details about training, license, limitations, etc. 
        \item The paper should discuss whether and how consent was obtained from people whose asset is used.
        \item At submission time, remember to anonymize your assets (if applicable). You can either create an anonymized URL or include an anonymized zip file.
    \end{itemize}

\item {\bf Crowdsourcing and Research with Human Subjects}
    \item[] Question: For crowdsourcing experiments and research with human subjects, does the paper include the full text of instructions given to participants and screenshots, if applicable, as well as details about compensation (if any)? 
    \item[] Answer: \answerNA{} 
    \item[] Justification: This paper does not involve crowdsourcing nor research with human subjects
    \item[] Guidelines:
    \begin{itemize}
        \item The answer NA means that the paper does not involve crowdsourcing nor research with human subjects.
        \item Including this information in the supplemental material is fine, but if the main contribution of the paper involves human subjects, then as much detail as possible should be included in the main paper. 
        \item According to the NeurIPS Code of Ethics, workers involved in data collection, curation, or other labor should be paid at least the minimum wage in the country of the data collector. 
    \end{itemize}

\item {\bf Institutional Review Board (IRB) Approvals or Equivalent for Research with Human Subjects}
    \item[] Question: Does the paper describe potential risks incurred by study participants, whether such risks were disclosed to the subjects, and whether Institutional Review Board (IRB) approvals (or an equivalent approval/review based on the requirements of your country or institution) were obtained?
    \item[] Answer: \answerNA{} 
    \item[] Justification: This paper does not involve crowdsourcing nor research with human subjects
    \item[] Guidelines:
    \begin{itemize}
        \item The answer NA means that the paper does not involve crowdsourcing nor research with human subjects.
        \item Depending on the country in which research is conducted, IRB approval (or equivalent) may be required for any human subjects research. If you obtained IRB approval, you should clearly state this in the paper. 
        \item We recognize that the procedures for this may vary significantly between institutions and locations, and we expect authors to adhere to the NeurIPS Code of Ethics and the guidelines for their institution. 
        \item For initial submissions, do not include any information that would break anonymity (if applicable), such as the institution conducting the review.
    \end{itemize}

\end{enumerate}
\end{document}